# A New Look at BDDs for Pseudo-Boolean Constraints

**Ignasi Abío** IABIO@LSI.UPC.EDU
**Robert Nieuwenhuis** ROBERTO@LSI.UPC.EDU
**Albert Oliveras** OLIVERAS@LSI.UPC.EDU
**Enric Rodríguez-Carbonell** ERODRI@LSI.UPC.EDU
*Technical University of Catalonia (UPC), Barcelona.*

**Valentin Mayer-Eichberger** MAYEREICHBERGER@GMAIL.COM

## Abstract

Pseudo-Boolean constraints are omnipresent in practical applications, and thus a significant effort has been devoted to the development of good SAT encoding techniques for them. Some of these encodings first construct a Binary Decision Diagram (BDD) for the constraint, and then encode the BDD into a propositional formula. These BDD-based approaches have some important advantages, such as not being dependent on the size of the coefficients, or being able to share the same BDD for representing many constraints. We first focus on the size of the resulting BDDs, which was considered to be an open problem in our research community. We report on previous work where it was proved that there are Pseudo-Boolean constraints for which no polynomial BDD exists. We also give an alternative and simpler proof assuming that NP is different from Co-NP. More interestingly, here we also show how to overcome the possible exponential blowup of BDDs by *coefficient decomposition*. This allows us to give the first polynomial generalized arc-consistent ROBDD-based encoding for Pseudo-Boolean constraints. Finally, we focus on practical issues: we show how to efficiently construct such ROBDDs, how to encode them into SAT with only 2 clauses per node, and present experimental results that confirm that our approach is competitive with other encodings and state-of-the-art Pseudo-Boolean solvers.

## 1. Introduction

In this paperwe study Pseudo-Boolean constraints (PB constraints for short), that is, constraints of the form $a_1x_1 + \cdots + a_nx_n \# K$, where the $a_i$ and $K$ are integer coefficients, the $x_i$ are Boolean (0/1) variables, and the relation operator $\#$ belongs to $\{<, >, \leq, \geq, =\}$. We will assume that $\#$ is $\leq$ and the $a_i$ and $K$ are positive since other cases can be easily reduced to this one (see Eén & Sörensson, 2006).

Such a constraint ($\leq$ with positive coefficients) is a Boolean function $C\colon \{0,1\}^n \to \{0,1\}$ that is monotonic decreasing in the sense that any solution for $C$ remains a solution after flipping inputs from 1 to 0. Therefore these constraints can be expressed by a set of clauses with only negative literals. For example, each clause could simply define a (minimal) subset of variables that cannot be simultaneously true. Note however that not every such a monotonic function is a PB constraint. For example, the function expressed by the two clauses $\overline{x}_1 \vee \overline{x}_2$ and $\overline{x}_3 \vee \overline{x}_4$ has no (single) equivalent PB constraint $a_1x_1 + \cdots + a_4x_4 \leq K$ (since without loss of generality $a_1 \geq a_2$ and $a_3 \geq a_4$, and then also $\overline{x}_1 \vee \overline{x}_3$ is needed). Hence, even among the monotonic Boolean functions, PB constraints are a rather restricted class (see also Smaus, 2007).

PB constraints are omnipresent in practical SAT applications, not just in typical 0-1 linear integer problems, but also as an ingredient in new SAT approaches to, e.g., cumulative scheduling (Schutt, Feydy, Stuckey, & Wallace, 2009), logic synthesis (Aloul, Ramani, Markov, & Sakallah, 2002) or verification (Bryant, Lahiri, & Seshia, 2002), so it is not surprising that a significant number of SAT encodings for these constraints have been proposed in the literature. Here we are interested in encoding a PB constraint $C$ by a clause set $S$ (possibly with auxiliary variables) that is not only equisatisfiable, but also *generalized arc-consistent* (GAC): given a partial assignment $A$, if $x_i$ is false in every extension of $A$ satisfying $C$, then unit propagating $A$ on $S$ sets $x_i$ to false.

To our knowledge, the only polynomial GAC encoding so far was given by Bailleux, Boufkhad, and Roussel (2009). Some other existing encodings are based on building (forms of) Binary Decision Diagrams (BDDs) and translating these into CNF. Although the approach of Bailleux et al. is not BDD-based, our main motivation to revisit BDD-based encodings is the following:

**Example 1.** *Let us consider two Pseudo-Boolean constraints:* $3x_1 + 2x_2 + 4x_3 \leq 5$ *and* $30001x_1 + 19999x_2 + 39998x_3 \leq 50007$*. Both are clearly equivalent: the Boolean function they represent can be expressed, e.g., by the clauses* $\overline{x}_1 \vee \overline{x}_3$ *and* $\overline{x}_2 \vee \overline{x}_3$*. However, encodings like the one of Bailleux et al. (2009) heavily depend on the concrete coefficients of each constraint, and generate a significantly larger SAT encoding for the second one. Since, given a variable ordering, ROBDDs are a canonical representation for Boolean functions (Bryant, 1986), i.e., each Boolean function has a unique ROBDD, a ROBDD-based encoding will treat both constraints equivalently.*

Another reason for revisiting BDDs is that in practical problems numerous PB constraints exist that share variables among each other. Representing them all as a single ROBDD has the potential of generating a much more compact SAT encoding that is moreover likely to have better propagation properties.

As we have mentioned, BDD-based approaches have already been studied in the literature. A good example is the work of Eén and Sörensson (2006), where a GAC encoding using six three-literals clauses per BDD node is given. However, when it comes to study the BDD size, on page 9 they cite the work of Bailleux, Boufkhad, and Roussel (2006) to say *"It is proven that in general a PB-constraint can generate an exponentially sized BDD"*. In Section 7 we explain why the approach of Bailleux et al does not use ROBDDs, and prove that the example they use to show the exponentiality of their method turns out to have polynomial ROBDDs. Somewhat surprisingly, probably due to the different names that PB constraints receive (0-1 integer linear constraints, linear threshold functions, weight constraints, knapsack constraints), the work of Hosaka, Takenaga, and Yajima (1994) has remained unknown to our research community. In that paper, it is proved that there are PB constraints for which no polynomial-sized ROBDDs exist. For self-containedness of this article, and to bring this interesting result to the knowledge of our research community, we include this family of PB constraints and prove that, regardless of the variable ordering, the corresponding ROBDD will always have exponential size.

**Main contributions and organization of this paper:**

- Subsection 3.2: We reproduce the family of PB constraints proposed by Hosaka et al. (1994), for which no polynomial-size ROBDD exist. For self-containedness, we give a clearer alternative proof than in the original paper.
- Subsection 3.3: A very simple proof that, unless NP=co-NP, there are PB constraints that admit no polynomial-size ROBDD, independently of the variable order.
- Subsection 4.1: A proof that PB constraints whose coefficients are powers of two do admit polynomial-size ROBDDs.
- Subsections 4.2 and 4.3: A GAC and polynomial (size $\mathcal{O}(n^3 \log a_{max})$) ROBDD-based encoding for PB constraints.
- Section 5: An algorithm to construct ROBDDs for Pseudo-Boolean constraints in polynomial time w.r.t. the size of the final ROBDD.
- Section 6: A GAC SAT encoding of BDDs for monotonic functions, a more general class of Boolean functions than PB constraints. This encoding uses only one binary and one ternary clause per node (the standard if-then-else encoding for BDDs used in, e.g., Eén & Sörensson, 2006, requires six ternary clauses per node). Moreover, this translation works for any BDD variable ordering.
- Section 7: A related work section, summarizing the most important ingredients of the existing encodings of Pseudo-Boolean constraints into SAT.
- Section 8: An experimental evaluation comparing our approach with other encodings and tools.

This article extends the shorter preliminary paper "BDDs for Pseudo-Boolean Constraints – Revisited" (Abío, Nieuwenhuis, Oliveras, & Rodríguez-Carbonell, 2011), which was presented at the SAT 2011 conference. Extensions include: (i) proofs of all technical results, (ii) multiple examples illustrating the various concepts and algorithms presented, (iii) the PB constraint family by Hosaka et al. (1994) for which no polynomial ROBDD exists, (iv) an algorithm to efficiently construct ROBDDs for Pseudo-Boolean constraints, (v) a detailed related work section, (vi) extensive experimental results comparing our encoding to other approaches and (vii) a brief report of our experience trying to take advantage of the sharing potential of BDDs.

## 2. Preliminaries

Let $\mathcal{X} = \{x_1, x_2, \ldots\}$ be a fixed set of propositional *variables*. If $x \in \mathcal{X}$ then $x$ and $\overline{x}$ are *positive and negative literals*, respectively. The *negation* of a literal $l$, written $\bar{l}$, denotes $\overline{x}$ if $l$ is $x$, and $x$ if $l$ is $\overline{x}$. A *clause* is a disjunction of literals $\overline{x}_1 \vee \ldots \vee \overline{x}_p \vee x_{p+1} \vee \ldots \vee x_n$, sometimes written as $x_1 \wedge \ldots x_p \rightarrow x_{p+1} \vee \ldots x_n$. A *CNF* formula is a conjunction of clauses.

A (partial) *assignment* $A$ is a set of literals such that $\{x, \overline{x}\} \subseteq A$ for no $x$, i.e., no contradictory literals appear. A literal $l$ is *true* in $A$ if $l \in A$, is *false* in $A$ if $\bar{l} \in A$, and is *undefined* in $A$ otherwise. Sometimes we will write $A$ as a set of pairs $x = v$, where $v$ is 1 if $x$ is true in $A$ and 0 if $x$ is false in $A$. A clause $C$ is true in $A$ if at least one of its literals is true in $A$. A formula $F$ is true in $A$ if all its clauses are true in $A$. In that case, $A$ is a *model* of $F$. Systems that decide whether a formula $F$ has any model are called SAT-solvers, and the main inference rule they implement is *unit propagation*: given a CNF $F$ and an assignment $A$, find a clause in $F$ such that all its literals are false in $A$ except one, say $l$, which is undefined, add $l$ to $A$ and repeat the process until reaching a fixpoint.

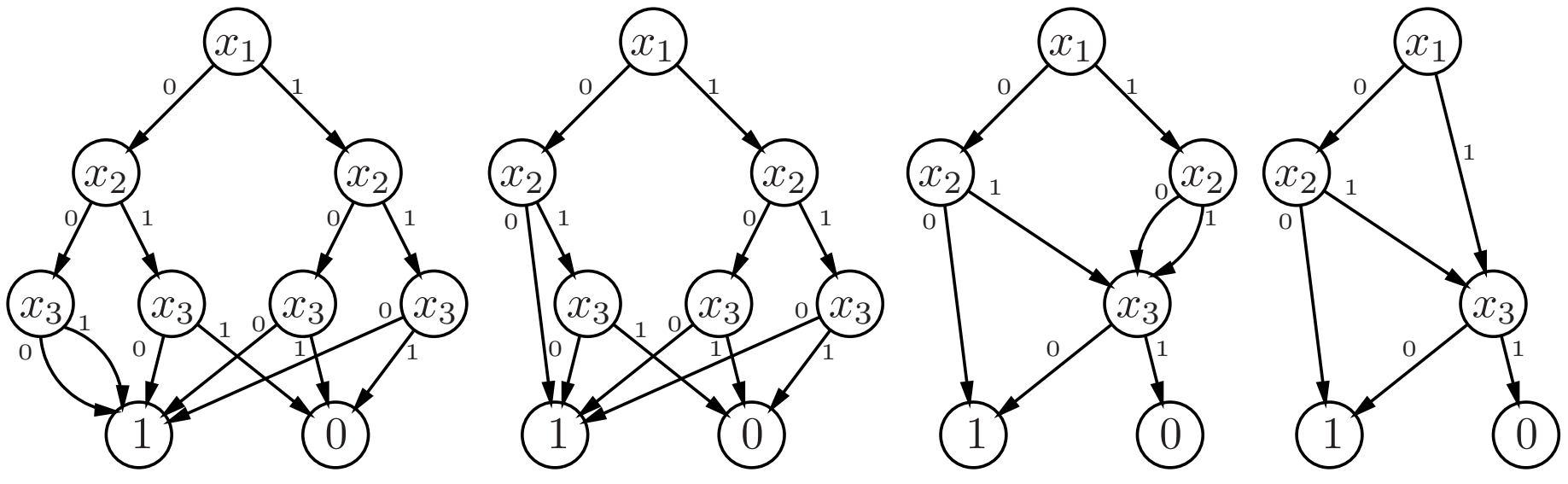

Figure 1: Construction of a BDD for $2x_1 + 3x_2 + 5x_3 \leq 6$

Pseudo-Boolean constraints (PB constraints for short) are constraints of the form $a_1x_1 + \cdots + a_nx_n \;\#\; K$, where the $a_i$ and $K$ are integer coefficients, the $x_i$ are Boolean (0/1) variables, and the relation operator $\#$ belongs to $\{<, >, \leq, \geq, =\}$. We will assume that $\#$ is $\leq$ and the $a_i$ and $K$ are positive, since other cases can be easily reduced to this one [1]: (i) changing into $\leq$ is straightforward if coefficients can be negative; (ii) replacing $-ax$ by $a(1-x)-a$; (iii) replacing $(1-x)$ by $\overline{x}$. Negated variables like $\overline{x}$ can be handled as positive ones or, alternatively, replaced by a fresh $x'$ and adding the clauses $x \vee x'$ and $\overline{x} \vee \overline{x}'$. A particular case of Pseudo-Boolean constraints is the one of *cardinality constraints*, in which all coefficients $a_i$ are equal to 1.

Our main goal is to find CNF encodings for PB constraints. That is, given a PB-constraint $C$, construct an equisatisfiable clause set (a CNF) $S$ such that any model for $S$ restricted to the variables of $C$ is a model of $C$ and viceversa. Two extra properties are sought: (i) *consistency checking by unit propagation* or simply *consistency*: whenever a partial assignment $A$ cannot be extended to a model for $C$, unit propagation on $S$ and $A$ produces a contradiction (a literal $l$ and its negation $\overline{l}$); and (ii) *generalized arc-consistency* or *GAC* (again by unit propagation): given an assignment $A$ that can be extended to a model of $C$, but such that $A \cup \{x\}$ cannot, unit propagation on $S$ and $A$ produces $\overline{x}$. More concretely, we will use ROBDDs for finding such encodings. ROBDDs are introduced by means of the following example.

**Example 2.** *Figure 1 explains (one method for) the construction of a ROBDD for the PB constraint* $2x_1 + 3x_2 + 5x_3 \leq 6$ *and the ordering* $[x_1, x_2, x_3]$. *The root node has as* selector variable $x_1$. *Its* false child *represents the PB constraint assuming* $x_1 = 0$ *(i.e.,* $3x_2 + 5x_3 \leq 6$*) and its* true child *represents* $2 + 3x_2 + 5x_3 \leq 6$*, that is,* $3x_2 + 5x_3 \leq 4$. *The two children have the next variable in the ordering* ($x_2$) *as selector, and the process is repeated until we reach*

1. An =-constraint can be split into a $\leq$-constraint and a $\geq$-constraint. Here we consider (generalized arc-)consistency for the latter two isolatedly, not for the original =-constraint.

*the last variable in the sequence. Then, a constraint of the form* $0 \leq K$ *is the* True *node (1 in the figure) if* $K \geq 0$ *is positive, and the* False *node (0) if* $K < 0$*. This construction (leftmost in the figure), is known as an Ordered BDD. For obtaining a Reduced Ordered BDD (ROBDD for short in the rest of the paper), two reductions are applied until fixpoint: removing nodes with identical children (as done with the leftmost* $x_3$ *node in the second BDD of the figure), and merging isomorphic subtrees, as done for* $x_3$ *in the third BDD. The fourth final BDD is a fixpoint. For a given ordering, ROBDDs are a canonical representation of Boolean functions: each Boolean function has a unique ROBDD. BDDs can be encoded into CNF by introducing an auxiliary variable* $a$ *for every node. If the selector variable of the node is* $x$ *and the auxiliary variables for the false and true child are* $f$ *and* $t$*, respectively, add the if-then-else clauses:*

$$\begin{array}{lll} \overline{x} \wedge \overline{f} \rightarrow \overline{a} & \quad x \wedge \overline{t} \rightarrow \overline{a} & \quad \overline{f} \wedge \overline{t} \rightarrow \overline{a} \\ \overline{x} \wedge f \rightarrow a & \quad x \wedge t \rightarrow a & \quad f \wedge t \rightarrow a \end{array}$$

In what follows, the *size* of a BDD is its number of nodes. We will say that a BDD *represents* a PB constraint if they represent the same Boolean function. Given an assignment $A$ over the variables of a BDD, we define the *path induced by* $A$ as the path that starts at the root of the BDD and at each step, moves to the false (true) child of a node if and only if its selector variable is false (true) in $A$.

## 3. Exponential ROBDDs for PB Constraints

In this section we study the size of ROBDDs for PB constraints. We start by defining the notion of the *interval* of a PB constraint. Then, in Section 3.2 we consider two families of PB constraints and study their ROBDD size: we first prove that the example given by Bailleux et al. (2006) has polynomial ROBDDs, and then we reproduce the example of Hosaka et al. (1994) that has exponential ROBDDs regardless of the variable ordering. Finally, we relate the ROBDD size of a PB constraint with the well-known subset sum problem.

### 3.1 Intervals

Before formally defining the notion of interval of a PB constraint, let us first give some intuitive explanation.

**Example 3.** *Consider the constraint* $2x_1 + 3x_2 + 5x_3 \leq 6$*. Since no combination of its coefficients adds to 6, the constraint is equivalent to* $2x_1 + 3x_2 + 5x_3 < 6$*, and hence to* $2x_1 + 3x_2 + 5x_3 \leq 5$*. This process cannot be repeated again since 5 can be obtained with the existing coefficients.*

*Similarly, we could try to increase the right-hand side of the constraint. However, there is a combination of the coefficients that adds to 7, which implies that the constraint is not equivalent to* $2x_1 + 3x_2 + 5x_3 \leq 7$*. All in all, we can state that the constraint is equivalent to* $2x_1 + 3x_2 + 5x_3 \leq K$ *for any* $K \in [5, 6]$*. It is trivial to see that the set of valid* $K$*'s is always an interval.*

**Definition 4.** *Let* $C$ *be a constraint of the form* $a_1x_1 + \cdots + a_nx_n \leq K$*. The* interval *of* $C$ *consists of all integers* $M$ *such that* $a_1x_1 + \cdots + a_nx_n \leq M$*, seen as a Boolean function, is equivalent to* $C$*.*

*Similarly, given a ROBDD representing a PB constraint and a node* $\nu$ *with selector variable* $x_i$*,we will refer to* the interval of $\nu$ *as all the integers* $M$ *such that the constraint* $a_i x_i + \cdots a_n x_n \le M$ *is represented (as a Boolean function) by the ROBDD rooted at* $\nu$*.*

In the following, unless stated otherwise, the ordering used in the ROBDD will be $[x_1, x_2, \ldots, x_n]$.

**Proposition 5.** *If* $[\beta, \gamma]$ *is the interval of a node* $\nu$ *with selector variable* $x_i$ *then:*

1. *There is an assignment* $\{x_j = v_j\}_{j=i}^{n}$ *such that* $a_i v_i + \cdots + a_n v_n = \beta$*.*
2. *There is an assignment* $\{x_j = v_j\}_{j=i}^{n}$ *such that* $a_i v_i + \cdots + a_n v_n = \gamma + 1$*.*
3. *There is an assignment* $\{x_j = v_j\}_{j=1}^{i-1}$ *such that* $K - a_1 v_1 - a_2 v_2 - \cdots - a_{i-1} v_{i-1} \in [\beta, \gamma]$
4. *Take* $h < \beta$*. There exists an assignment* $\{x_j = v_j\}_{j=i}^{n}$ *such that* $a_i v_i + \cdots + a_n v_n > h$ *and its path goes from* $\nu$ *to True.*
5. *Take* $h > \gamma$*. There exists an assignment* $\{x_j = v_j\}_{j=i}^{n}$ *such that* $a_i v_i + \cdots + a_n v_n \le h$ *and its path goes from* $\nu$ *to False.*
6. *The interval of the True node is* $[0, \infty)$*.*
7. *The interval of the False node is* $(-\infty, -1]$*. Moreover, it is the only interval with negative values.*

*Proof.*

1. Since $\beta - 1$ does not belong to the interval of $\nu$, the constraints

$$\begin{aligned} a_i x_i + a_{i+1} x_{i+1} + \cdots + a_n x_n &\le \beta - 1 \\ a_i x_i + a_{i+1} x_{i+1} + \cdots + a_n x_n &\le \beta \end{aligned}$$

   are different. This means that there is a partial assignment satisfying the second one but not the first one.

2. The proof is analogous to the previous one.

3. Take a partial assignment $\{x_1 = v_1, \ldots, x_{i-1} = v_{i-1}\}$ whose path goes from the root to $\nu$. Therefore, by definition of the ROBDD, $\nu$ is the ROBDD of the constraint

$$a_i x_i + a_{i+1} x_{i+1} + \cdots + a_n x_n \le K - a_1 v_1 - \cdots - a_{i-1} v_{i-1}.$$

   Therefore, by definition of the interval of $\nu$,

$$K - a_1 v_1 - a_2 v_2 - \cdots - a_{i-1} v_{i-1} \in [\beta, \gamma].$$

4. Intuitively, this property states that, if $h$ is not in the interval of $\nu$, there is an assignment that satisfies the ROBDD rooted at $\nu$ but not the constraint $a_i x_i + \cdots + a_n x_n \le h$.

   Since $h$ does not belong to the interval of $\nu$, the ROBDD of

$$C' : a_i x_i + \cdots + a_n x_n \le h$$

   is not $\nu$. Therefore, there exists an assignment that either

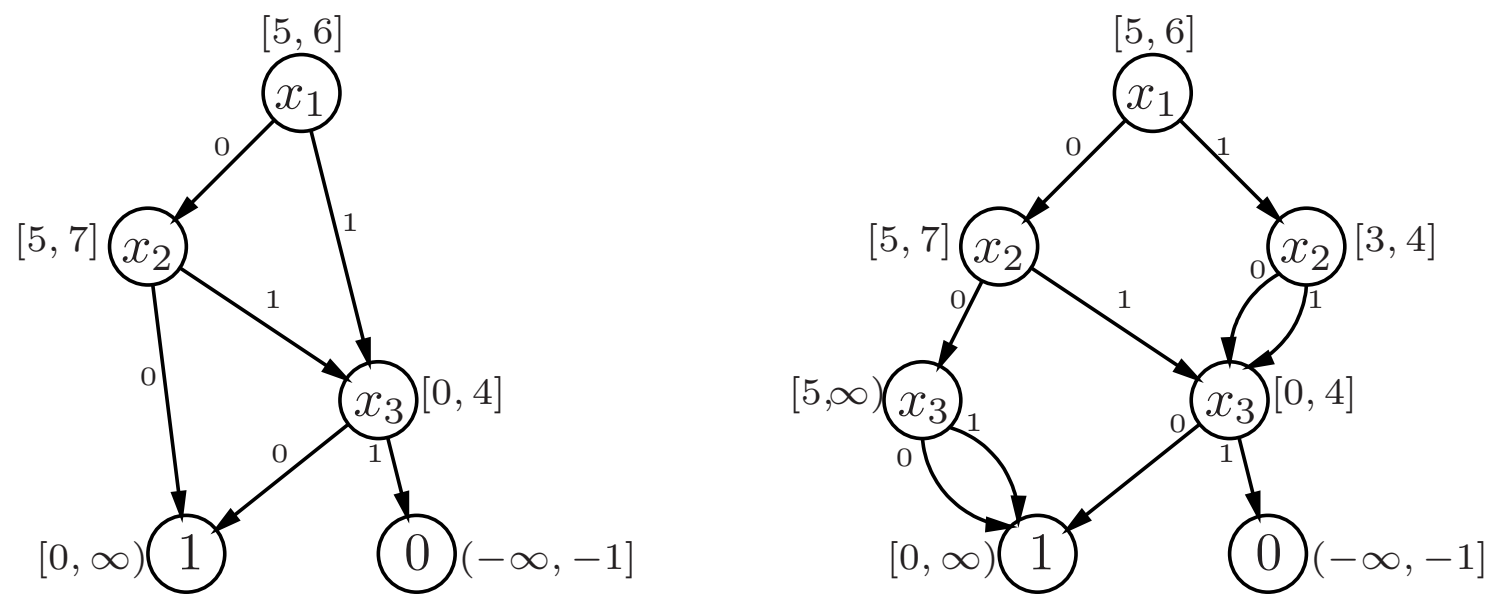

Figure 2: Intervals of the ROBDD for $2x_1 + 3x_2 + 5x_3 \leq 6$

(i) goes from $\nu$ to False but satisfies $C'$; or

(ii) goes to True but does not satisfy $C'$.

We want to prove that the assignment satisfies (ii). Assume that it satisfies (i). Since it goes from $\nu$ to False and $\beta$ belongs to the interval of $\nu$, it holds

$$a_i v_i + \cdots + a_n v_n > \beta.$$

Since $\beta > h$, the assignment does not satisfy $C'$, which is a contradiction. Therefore, the assignment satisfies (ii).

5. Take the assignment of the second point of this proposition. Since $\gamma + 1$ does not belong to the interval, the path of the assignment goes from $\nu$ to False. Moreover, $a_i v_i + \cdots + a_n v_n = \gamma + 1 \leq h$.

6. The True node is the ROBDD of the tautology. Therefore, it represents the PB constraint $0 \leq h$ for $h \in [0, \infty)$.

7. The False node is the ROBDD of the contradiction. Therefore, represents the PB constraint $0 \leq h$ for $h \in (-\infty, -1]$. Moreover, $a_i x_i + \cdots + a_n x_n < 0$ is also a contradiction, hence that constraint is also represented by the False node. Therefore, there is no other node with an interval with negative values.

□

We now prove that, given a ROBDD for a PB constraint, one can easily compute the intervals for every node bottom-up. We first start with a motivating example.

**Example 6.** *Let us consider again the constraint $2x_1 + 3x_2 + 5x_3 \leq 6$. Assume that all variables appear in every path from the root to the leaves (otherwise, add extra nodes as in the rightmost BDD of Figure 2). Assume now that we have computed the intervals for the two children of the root (rightmost BDD in Figure 2). This means that the false child of the root is the BDD for $3x_2+5x_3 \leq [5, 7]$ and the true child the BDD for $3x_2+5x_3 \leq [3, 4]$. Assuming $x_1$ to be false, the false child would also represent the constraint $2x_1+3x_2+5x_3 \leq [5, 7]$, and assuming $x_1$ to be true, the true child would represent the constraint $2x_1+3x_2+5x_3 \leq [5, 6]$. Taking the intersection of the two intervals, we can infer that the root node represents $2x_1 + 3x_2 + 5x_3 \leq [5, 6]$.*

More formally, the interval of every node can be computed as follows:

**Proposition 7.** *Let $a_1x_1 + a_2x_2 + \cdots + a_nx_n \le K$ be a constraint, and let $\mathcal{B}$ be its ROBDD with the order $[x_1, \ldots, x_n]$. Consider a node $\nu$ with selector variable $x_i$, false child $\nu_f$ (with selector variable $x_f$ and interval $[\beta_f, \gamma_f]$) and true child $\nu_t$ (with selector variable $x_t$ and interval $[\beta_t, \gamma_t]$), as shown in Figure 3. The interval of $\nu$ is $[\beta, \gamma]$, with:*

$$\beta = \max\{\beta_f + a_{i+1} + \cdots + a_{f-1},\ \ \beta_t + a_i + a_{i+1} + \cdots + a_{t-1}\},$$
$$\gamma = \min\{\gamma_f,\ \ \gamma_t + a_i\}.$$

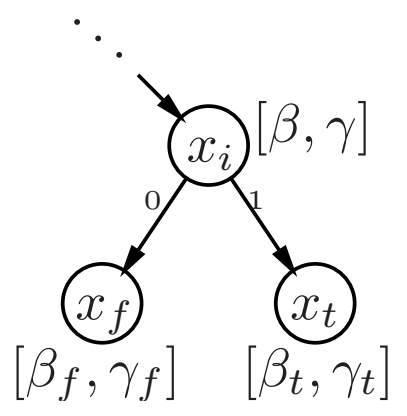

Figure 3: The interval of a node can be computed from its children's intervals.

Before moving to the proof, we want to note that if in every path from the root to the leaves of the ROBDD all variables were present, the definition of $\beta$ would be much simpler ($\beta = \max\{\beta_f, \beta_t + a_i\}$). The other coefficients are necessary to account for the variables that have been removed due to the ROBDD reduction process.

*Proof.* Let us assume that $[\beta, \gamma]$ is not the interval of $\nu$. One of the following statements should hold:

1. There exists $h \in [\beta, \gamma]$ that does not belong to the interval of $\nu$.

2. There exists $h < \beta$ belonging to the interval of $\nu$.

3. There exists $h > \gamma$ belonging to the interval of $\nu$.

We will now prove that none of these cases can hold.

1. Let us define
$$C' : a_ix_i + \cdots + a_nx_n \le h.$$
If $h$ does not belong to the interval, there exists an assignment $\{x_j = v_j\}_{j=i}^{n}$ that either satisfies $C'$ and its path goes from $\nu$ to False or it does not satisfy $C'$ and its path goes to True. Assume that the assignment satisfies $C'$ and its path goes from $\nu$ to False (the other case is similar). There are two possibilities:

   - The assignment satisfies $v_i = 0$. Since $h \ge \beta$, it holds
$$\begin{array}{rcl} h - a_{i+1}v_{i+1} - \cdots - a_{f-1}v_{f-1} & \ge & \beta - a_{i+1}v_{i+1} - \cdots - a_{f-1}v_{f-1} \\ & \ge & \beta - a_{i+1} - \cdots - a_{f-1} \ge \beta_f. \end{array}$$

On the other hand, since $h \leq \gamma$,

$$h - a_{i+1}v_{i+1} - \cdots - a_{f-1}v_{f-1} \leq h \leq \gamma \leq \gamma_f.$$

Therefore, $h - a_{i+1}v_{i+1} - \cdots - a_{f-1}v_{f-1}$ belongs to the interval of $\nu_f$. Since the assignment $\{x_f = v_f, \ldots, x_n = v_n\}$ goes from $\nu_f$ to False, we have:

$$a_f v_f + \cdots + a_n v_n > h - a_{i+1}v_{i+1} - \cdots - a_{f-1}v_{f-1}$$

$$a_{i+1}v_{i+1} + \cdots + a_f v_f + \cdots a_n v_n > h$$

Hence, adding $a_i v_i$ to the sum one can see that the assignment does not satisfy $C'$, which is a contradiction.

- The case $v_i = 1$ gives a similar contradiction.

2. By definition of $\beta$, either $h < \beta_f + a_{i+1} + \cdots + a_{f-1}$ or $h < \beta_t + a_i + a_{i+1} + \cdots + a_{t-1}$. We will only consider the first case, since the other one is similar. Therefore, $h - a_{i+1} - \cdots - a_{f-1} < \beta_f$. Due to point 4 of Proposition 5, there exists an assignment $\{x_f = v_f, \ldots x_n = v_n\}$ such that

$$a_f v_f + \cdots a_n v_n > h - a_{i+1} - \cdots - a_{f-1}$$

and its path goes from $\nu_f$ to True. Hence, the assignment

$$\{x_i = 0, x_{i+1} = 1, \ldots, x_{f-1} = 1, x_f = v_f, \ldots, x_n = v_n\}$$

does not satisfy the constraint $a_i x_i + \cdots + a_n x_n \leq h$ and its path goes from $\nu$ to True. By definition of interval, $h$ cannot belong to the interval of $\nu$.

3. This case is very similar to the previous one.

□

This proposition gives a natural way of computing all intervals of a ROBDD in a bottom-up fashion. The procedure is initialized by computing the intervals of the terminal nodes as detailed in Proposition 5, points 6 and 7.

**Example 8.** *Let us consider again the constraint* $2x_1 + 3x_2 + 5x_3 \leq 6$. *Its ROBDD is shown in the left-hand side of Figure 2, together with its intervals. For their computation, we first compute the intervals of the True and False nodes, which are* $[0, \infty)$ *and* $(-\infty, -1]$ *in virtue of Proposition 5. Then, we can compute the interval of the node having* $x_3$ *as selector variable with the previous proposition's formula:* $\beta_3 = \max\{0, -\infty + 5\} = 0$, $\gamma_3 = \min\{\infty, -1 + 5\} = 4$. *Therefore, its interval is* $[0, 4]$.

*In the next step, we compute the interval for the node with selector variable* $x_2$*:* $\beta_2 = \max\{0 + 5, 0 + 3\} = 5$, $\gamma_2 = \min\{\infty, 4 + 3\} = 7$. *Thus, it its interval is* $[5, 7]$. *Finally, we can compute the root's interval:* $\beta_1 = \max\{5, 0 + 2 + 3\} = 5$, $\gamma_1 = \min\{7, 4 + 2\} = 6$, *that is,* $[5, 6]$.

### 3.2 Some Families of PB Constraints and their ROBDD Size

We start by revisiting the family of PB constraints given by Bailleux et al. (2006), where it is proved that, for their concrete variable ordering, their non-reduced BDDs grow exponentially for this family. Here we prove that ROBDDs are polynomial for this family, and that this is even independent of the variable ordering. The family is defined by considering $a$, $b$ and $n$ positive integers such that $\sum_{i=1}^{n} b^i < a$. The coefficients are $\omega_i = a + b^i$ and the right-hand side of the constraint is $K = a \cdot n/2$. We will first prove that the constraint $C : \omega_1 x_1 + \cdots + \omega_n x_n \leq K$ is equivalent to the cardinality constraint $C' : x_1 + \cdots + x_n \leq n/2 - 1$. For simplicity, we assume that $n$ is even.

- Take an assignment satisfying $C'$. In this case, there are at most $n/2 - 1$ variables $x_i$ assigned to true, and the assignment also satisfies $C$ since:

$$\omega_1 x_1 + \cdots + \omega_n x_n \leq \sum_{i=n/2+2}^{n} \omega_i = (n/2-1)a + \sum_{i=n/2+2}^{n} b^i < K - a + \sum_{i=1}^{n} b^i < K.$$

- Consider now an assignment not satisfying $C'$. In this case, there are at least $n/2$ true variables in the assignment and it does not satisfy $C$ either:

$$\omega_1 x_1 + \cdots + \omega_n x_n \geq \sum_{i=1}^{n/2} \omega_i = (n/2) \cdot a + \sum_{i=1}^{n/2} b^i > (n/2) \cdot a = K.$$

Since the two constraints are equivalent and ROBDDs are canonical, the ROBDD representation of $C$ and $C'$ are the same. But the ROBDD of $C'$ is known to be of quadratic size because it is a cardinality constraint (see, for instance, Bailleux et al., 2006).

In the following, we present a family of PB constraints that only admit exponential ROBDDs. This example was first given by Hosaka et al. (1994), but a clearer alternative proof is given next. First of all, we prove a lemma that, under certain technical conditions, gives a lower bound on the number of nodes of the ROBDD for a PB constraint.

**Lemma 9.** *Let $a_1 x_1 + \cdots + a_n x_n \leq K$ be a PB constraint, and let $i$ be an integer with $1 \leq i \leq n$. Assume that every assignment $\{x_1 = v_1, x_2 = v_2, \ldots, x_i = v_i\}$ admits an extension $\{x_1 = v_1, \ldots, x_n = v_n\}$ such that $a_1 v_1 + \cdots + a_n v_n = K$. Let $M$ be the number of different results we can obtain adding some subset of the coefficients $a_1, a_2, \ldots, a_i$, i.e., $M = |\{\sum_{j=1}^{i} a_j b_j \ : \ b_j \in \{0,1\}\}|$. Then, the ROBDD size with ordering $[x_1, x_2, \ldots, x_n]$ is at least $M$.*

*Proof.* Let us consider a PB constraint that satisfies the conditions of the lemma. We will prove that its ROBDD has at least $M$ distinct nodes by showing that any two assignments of the form $\{x_1 = v_1, \ldots, x_i = v_i\}$ and $\{x_1 = v'_1, \ldots, x_i = v'_i\}$ with $a_1 v_1 + \cdots + a_i v_i \neq a_1 v'_1 + \cdots + a_i v'_i$ lead to different nodes in the ROBDD.

Assume that it is not true: there are two assignments $\{x_1 = v_1, \ldots, x_i = v_i\}$ and $\{x_1 = v'_1, \ldots, x_i = v'_i\}$ with $a_1 v_1 + \cdots + a_i v_i < a_1 v'_1 + \cdots + a_i v'_i$ such that their paths end at the same node. Take the extended assignment $A = \{x_1 = v_1, \ldots, x_n = v_n\}$ such that $a_1 v_1 + \cdots a_n v_n = K$. Since $A$ satisfies the PB constraint, the path it defines ends at the true node. However, the assignment $A' = \{x_1 = v'_1, \ldots, x_i = v'_i, x_{i+1} = v_{i+1}, \ldots, x_n = v_n\}$ does not satisfy the constraint, since

$$a_1 v'_1 + \cdots a_i v'_i + a_{i+1} v_{i+1} + \cdots a_n v_n > a_1 v_1 + \cdots + a_n v_n = K.$$

However, the nodes defined by $\{x_1 = v_1, \ldots, x_i = v_i\}$ and $\{x_1 = v'_1, \ldots, x_i = v'_i\}$ were the same, so the path defined by $A'$ must also end at the true node, which is a contradiction. $\square$

We can now show a family of PB constraints that only admits exponential ROBDDs.

**Theorem 10.** *Let $n$ be a positive integer, and let us define $a_{i,j} = 2^{j-1} + 2^{2n+i-1}$ for all $1 \le i, j \le 2n$; and $K = (2^{4n} - 1)n$. Then, the PB constraint*

$$\sum_{i=1}^{2n} \sum_{j=1}^{2n} a_{i,j} x_{i,j} \le K$$

*has at least $2^n$ nodes in any variable ordering.*

*Proof.* It is convenient to describe the coefficients in binary notation:

$$
\begin{array}{rcccccccccccc}
 & & \multicolumn{5}{c}{\overbrace{\hphantom{0\quad 0\quad \cdots\quad 0\quad 1}}^{2n}} & & \multicolumn{5}{c}{\overbrace{\hphantom{0\quad 0\quad \cdots\quad 0\quad 1}}^{2n}} \\
a_{1,1} & = & 0 & 0 & \cdots & 0 & \mathbf{1} & & 0 & 0 & \cdots & 0 & \mathbf{1} \\
a_{1,2} & = & 0 & 0 & \cdots & 0 & \mathbf{1} & & 0 & 0 & \cdots & \mathbf{1} & 0 \\
\cdots & & & & & & & & & & \iddots & & \\
a_{1,2n-1} & = & 0 & 0 & \cdots & 0 & \mathbf{1} & & 0 & \mathbf{1} & \cdots & 0 & 0 \\
a_{1,2n} & = & 0 & 0 & \cdots & 0 & \mathbf{1} & & \mathbf{1} & 0 & \cdots & 0 & 0 \\
 & & & & & & & & & & & & \\
a_{2,1} & = & 0 & 0 & \cdots & \mathbf{1} & 0 & & 0 & 0 & \cdots & 0 & \mathbf{1} \\
a_{2,2} & = & 0 & 0 & \cdots & \mathbf{1} & 0 & & 0 & 0 & \cdots & \mathbf{1} & 0 \\
\cdots & & & & & & & & & & \iddots & & \\
a_{2,2n-1} & = & 0 & 0 & \cdots & \mathbf{1} & 0 & & 0 & \mathbf{1} & \cdots & 0 & 0 \\
a_{2,2n} & = & 0 & 0 & \cdots & \mathbf{1} & 0 & & \mathbf{1} & 0 & \cdots & 0 & 0 \\
\cdots & & & & \iddots & & & & & & & & \\
 & & & & & & & & & & & & \\
a_{2n,2n} & = & \mathbf{1} & 0 & \cdots & 0 & 0 & & \mathbf{1} & 0 & \cdots & 0 & 0 \\
 & & & & & & & & & & & & \\
K/n & = & \mathbf{1} & \mathbf{1} & \cdots & \mathbf{1} & \mathbf{1} & & \mathbf{1} & \mathbf{1} & \cdots & \mathbf{1} & \mathbf{1}
\end{array}
$$

First of all, one can see that the sum of all the $a$'s is $2K$.

Let us take an arbitrary bijection

$$F = (F_1, F_2) : \{1, 2, \ldots, 4n^2\} \to \{1, 2, \ldots, 2n\} \times \{1, 2, \ldots, 2n\},$$

and consider the ordering defined by it: $[x_{F(1)}, x_{F(2)}, \ldots, x_{F(4n^2)}]$, where $x_{F(k)} = x_{F_1(k), F_2(k)}$ for every $k$. We want to prove that the ROBDD of the PB constraint with this ordering has at least $2^n$ nodes.

The proof will consist in showing that the hypotheses of Lemma 9 hold. That is, first we show that for any variable ordering, we can find an integer $s$ such that any assignment to the first $s$ variables can be extended to a full assignment that adds $K$. Then, we prove that there are at least $2^n$ different values we can add with the first $s$ coefficients, as required by Lemma 9.

Let us define $b_k$ with $1 \leq k \leq 2n$ as the position of the $k$-th different value of the tuple $(F_1(1), F_1(2), \dots, F_1(4n^2))$. More formally,

$$b_k = \begin{cases} 1 & \text{if } k = 1, \\ \min\Big\{ r \ : \ F_1(r) \notin \{F_1(b_1), F_1(b_2), \dots, F_1(b_{k-1})\} \Big\} & \text{if } k > 1. \end{cases}$$

Analogously, let us define $c_1, \dots, c_{2n}$ as

$$c_k = \begin{cases} 1 & \text{if } k = 1, \\ \min\Big\{ s \ : \ F_2(s) \notin \{F_2(c_1), F_2(c_2), \dots, F_2(c_{k-1})\} \Big\} & \text{if } k > 1. \end{cases}$$

Let us denote by $i_r = F_1(b_r)$ and $j_s = F_2(c_s)$ for all $1 \leq r, s \leq 2n$. Notice that $\{i_1, i_2, \dots, i_{2n}\}$ and $\{j_1, j_2, \dots, j_{2n}\}$ are permutations of $\{1, 2, \dots, 2n\}$. Assume that $b_n \geq c_n$ (the other case is analogous), and take an arbitrary assignment $\{x_{F(1)} = v_{F(1)}, x_{F(2)} = v_{F(2)}, \dots, x_{F(c_n)} = v_{F(c_n)}\}$. We want to extend it to a complete assignment such that

$$\sum_{k=1}^{4n^2} a_{F(k)} v_{F(k)} = K.$$

Figure 4 represents the initial assignment. All the values are in the top-left square since the assignment is undefined for all $x_{i_r, j_s}$ with $r > n$ or $s > n$. Extending the assignment so that the sum is $K$ amounts to completing the table in such a way that there are exactly $n$ ones in every column and row.

The assignment can be completed in the following way: first, complete the top left square in any way, for instance, adding zeros to every non-defined cell. Then, copy that square to the bottom-right square and, finally, add the complementary square to the other two squares (i.e., write 0 instead of 1 and 1 instead of 0). Figure 5 shows the extended assignment for that example.

More formally, the assignment is completed as follows:

$$v_{i_r, j_s} = \begin{cases} 0 & \text{if } r, s \leq n \text{ and } v_{i_r, j_s} \text{ was undefined}, \\ \neg v_{i_{r-n}, j_s} & \text{if } r > n \text{ and } s \leq n, \\ \neg v_{i_r, j_{s-n}} & \text{if } s > n \text{ and } r \leq n, \\ v_{i_{r-n}, j_{s-n}} & \text{if } r, s > n, \end{cases}$$

where $\neg 0 = 1$ and $\neg 1 = 0$.

Now, let us prove that it satisfies the requirements, i.e., the coefficients corresponding to true variables in the assignment add exactly $K$. Let us fix $r, s \leq n$. Denote by $i = i_r$, $j = j_s$, $i' = i_{r+n}$ and $j' = j_{s+n}$.

- If $v_{i,j} = 0$, by definition $v_{i',j} = v_{i,j'} = 1$ and $v_{i',j'} = 0$. Therefore,

$$\begin{aligned} a_{i,j} v_{i,j} + a_{i',j} v_{i',j} + a_{i,j'} v_{i,j'} + a_{i',j'} v_{i',j'} &= a_{i',j} + a_{i,j'} \\ &= 2^{2n+i'-1} + 2^{j-1} + 2^{2n+i-1} + 2^{j'-1} \\ &= \frac{a_{i,j} + a_{i',j} + a_{i,j'} + a_{i',j'}}{2}. \end{aligned}$$

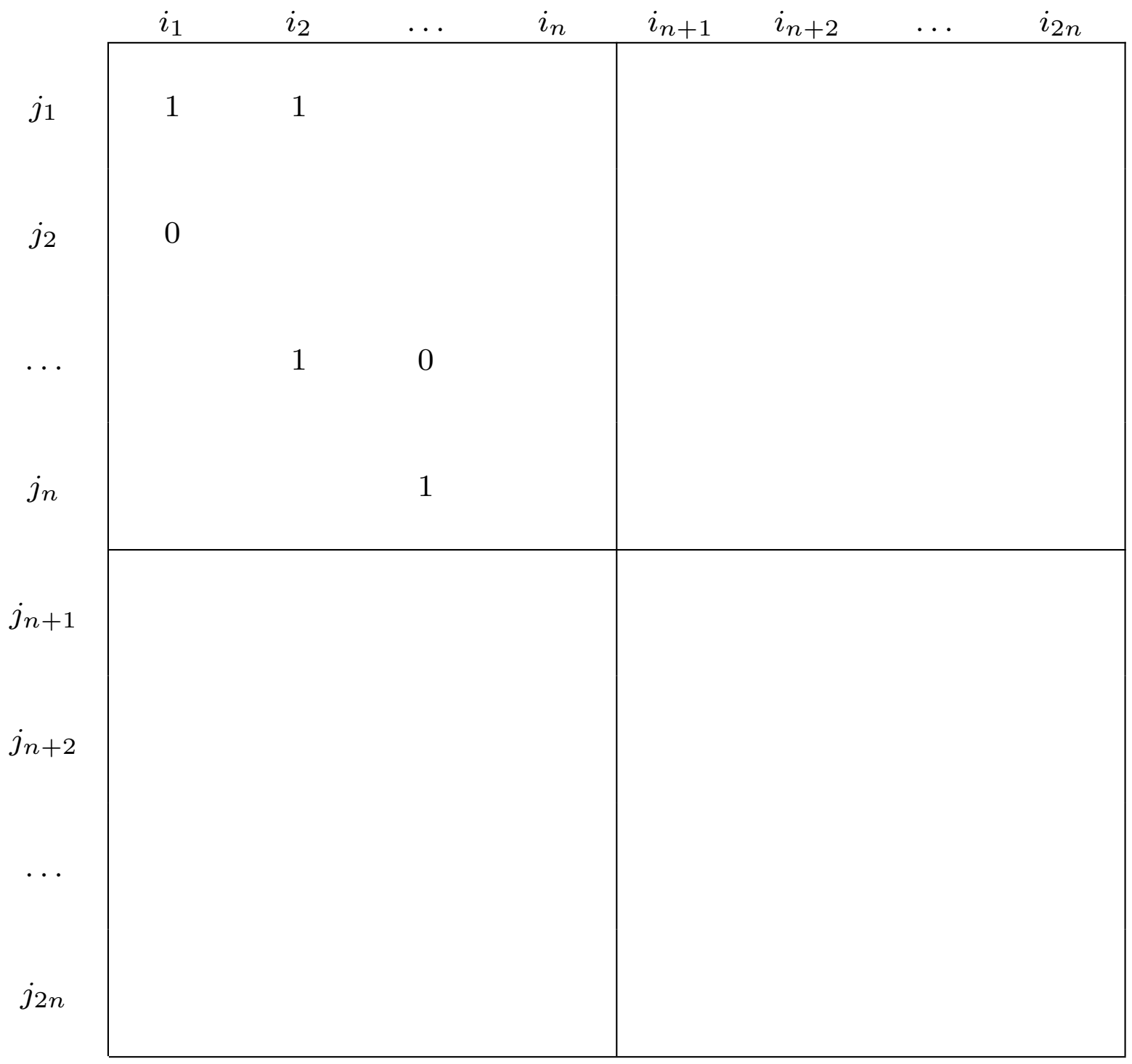

Figure 4: An arbitrary assignment. There is a 0, 1 or nothing at position $(i_r, j_s)$ depending on whether $x_{i_r,j_s}$ is false, true or unassigned.

| | $i_1$ | $i_2$ | ... | $i_n$ | $i_{n+1}$ | $i_{n+2}$ | ... | $i_{2n}$ |
|---|---|---|---|---|---|---|---|---|
| $j_1$ | **1** | **1** | 0 | 0 | 0 | 0 | 1 | 1 |
| $j_2$ | **0** | 0 | 0 | 0 | 1 | 1 | 1 | 1 |
| ... | 0 | **1** | **0** | 0 | 1 | 0 | 1 | 1 |
| $j_n$ | 0 | 0 | **1** | 0 | 1 | 1 | 0 | 1 |
| $j_{n+1}$ | 0 | 0 | 1 | 1 | 1 | 1 | 0 | 0 |
| $j_{n+2}$ | 1 | 1 | 1 | 1 | 0 | 0 | 0 | 0 |
| ... | 1 | 0 | 1 | 1 | 0 | 1 | 0 | 0 |
| $j_{2n}$ | 1 | 1 | 0 | 1 | 0 | 0 | 1 | 0 |

Figure 5: Extended assignment. There are exactly $n$ ones in every column and row.

- Analogously, if $v_{i,j} = 1$,

$$a_{i,j}v_{i,j} + a_{i',j}v_{i',j} + a_{i,j'}v_{i,j'} + a_{i',j'}v_{i',j'} = \frac{a_{i,j} + a_{i',j} + a_{i,j'} + a_{i',j'}}{2}.$$

Therefore,

$$\sum_{k=1}^{4n^2} a_{F(k)}v_{F(k)} = \frac{1}{2}\sum_{k=1}^{4n^2} a_{F(k)} = K.$$

By Lemma 9, the number of nodes of the ROBDD is at least the number of different results we can obtain by adding some subset of the coefficients $a_{F(1)}, a_{F(2)}, \dots, a_{F(c_n)}$. Consider the set $a_{F(c_1)}, a_{F(c_2)}, \dots, a_{F(c_n)}$. We will now see that all its different subsets add different values, and hence the ROBDD size is at least $2^n$.

The sum of a subset of $\{a_{F(c_1)}, a_{F(c_2)}, \dots, a_{F(c_n)}\}$ is

$$S = a_{F(c_1)}v_1 + a_{F(c_2)}v_2 + \cdots + a_{F(c_n)}v_n, \quad v_r \in \{0, 1\}.$$

Let us look at the $2n$ last bits of $S$ in binary notation: all the digits are 0 except for the positions $F_2(c_1), F_2(c_2), \dots, F_2(c_n)$, which are $v_1, v_2, \dots, v_n$. Therefore, if two subsets add the same, the $2n$ last digits of the sum are the same. This means that the values of $(v_1, \dots, v_n)$ are the same, and thus they are the same subset. □

### 3.3 Relation between the Subset Sum Problem and the ROBDD Size

In this section, we study the relationship between the ROBDD size for a PB constraint and the subset sum problem. This allows us to, assuming that NP and co-NP are different, give a much simpler proof that there exist PB constraints that do not admit polynomial ROBDDs.

Lemma 9 and the exponential ROBDD example of Theorem 10 suggest that there is a relationship between the size of ROBDDs and the number of ways we can obtain $K$ by adding some of the coefficients of the PB. It seems that if $K$ can be obtained in *a lot* of different ways, the ROBDD will be *large*.

In this section we explore another relationship between the problem of adding $K$ with a subset of the coefficients and the size of the ROBDDs. In a sense, we give a proof that the converse of the last paragraph is not true: if NP and co-NP are different, there are exponentially-sized ROBDDs of PB constraints with no subsets of their coefficients adding $K$. Let us start by defining one version of the well-known *subset sum* problem.

**Definition 11.** *Given a set of positive integers $S = \{a_1, \ldots, a_n\}$ and an integer $K$, the* subset sum problem *of $(K, S)$ consists in determining whether there exists a subset of $\{a_1, \ldots, a_n\}$ that sums to exactly $K$.*

It is well-known that the subset sum problem is NP-complete when $K \sim 2^n$, but there are polynomial algorithms in $n$ when $K$ is also a polynomial in $n$. For a given subset sum problem $(K, S)$ with $S = \{a_1, \ldots, a_n\}$, we can construct its *associated PB constraint* $a_1x_1 + \cdots + a_nx_n \le K$. In the previous section we have seen one PB constraint family whose coefficients can add $K$ in an exponential number of ways, thus generating an exponential ROBDD. Now, assuming that NP and co-NP are different, we will see that there exists a PB constraint family with exponential ROBDDs in any ordering such that their coefficients cannot add $K$. First, we show how ROBDDs can act as unsatisfiability certificates for the subset sum problem.

**Theorem 12.** *Let $(K, S)$ be an UNSAT subset sum problem. Then, if a ROBDD for its associated PB constraint has polynomial size, it can act as a polynomial unsatisfiability certificate for $(K, S)$.*

*Proof.* We only need to show how, in polynomial time, we can check whether the ROBDD is an unsatisfiability certificate for $(K, S)$. For that, we note that the subset sum problem is UNSAT if and only if the PB constraints

$$a_1x_1 + \cdots + a_nx_n \le K, \quad a_1x_1 + \cdots + a_nx_n \le K - 1$$

are equivalent, and this happens if and only if their ROBDDs are the same. Therefore, we have to show, in polynomial time, that the given ROBDD represents both constraints. It can be done, for instance, by building the ROBDD (using Algorithm 1 of Section 5) and comparing the ROBDDs. □

The key point now is that, if we assume NP to be different from co-NP, there exists a family of UNSAT subset sum problems with no polynomial-sized unsatisfiability certificate. Hence, the family consisting of the associated PB constraints does not admit polynomial ROBDDs.

Hence, PB constraints associated with difficult-to-certify UNSAT subset sum problems will produce exponential ROBDDs. However, subset sum is NP-complete if $K \sim 2^n$. In PB constraints from industrial problems usually $K \sim n^r$ for some $r$, so we could expect non-exponential ROBDDs for these constraints.

## 4. Avoiding Exponential ROBDDs

In this section we introduce our positive results. We restrict ourselves to a particular class of PB constraints, where all coefficients are powers of two. As we will show below, these constraints admit polynomial ROBDDs. Moreover, any PB constraint can be reduced to this class by means of *coefficient decomposition*.

**Example 13.** *Let us take the PB constraint* $9x_1 + 8x_2 + 3x_3 \leq 10$. *Considering the binary representation of the coefficients, this constraint can be rewritten into* $(2^3 x_{3,1} + 2^0 x_{0,1}) + (2^3 x_{3,2}) + (2^1 x_{1,3} + 2^0 x_{0,3}) \leq 10$ *if we add the binary clauses expressing that* $x_{i,r} = x_r$ *for appropriate* $i$ *and* $r$.

### 4.1 Power-of-two PB Constraints Do Have Polynomial-size ROBDDs

Let us consider a PB constraint of the form:

$$\begin{array}{rccccccccc} C: & 2^0 \cdot \delta_{0,1} \cdot x_{0,1} & + & 2^0 \cdot \delta_{0,2} \cdot x_{0,2} & + & \cdots & + & 2^0 \cdot \delta_{0,n} \cdot x_{0,n} & + & \\ & 2^1 \cdot \delta_{1,1} \cdot x_{1,1} & + & 2^1 \cdot \delta_{1,2} \cdot x_{1,2} & + & \cdots & + & 2^1 \cdot \delta_{1,n} \cdot x_{1,n} & + & \\ & & & & & \cdots & & & + & \\ & 2^m \cdot \delta_{m,1} \cdot x_{m,1} & + & 2^m \cdot \delta_{m,2} \cdot x_{m,2} & + & \cdots & + & 2^m \cdot \delta_{m,n} \cdot x_{m,n} & \leq & K, \end{array}$$

where $\delta_{i,r} \in \{0,1\}$ for all $i$ and $r$. Notice that every PB constraint whose coefficients are powers of 2 can be expressed in this way. Let us consider its ROBDD representation with the ordering $[x_{0,1}, x_{0,2}, \ldots, x_{0,n}, x_{1,1}, \ldots, x_{m,n}]$.

**Lemma 14.** *Let* $[\beta, \gamma]$ *be the interval of a node with selector variable* $x_{i,r}$. *Then* $2^i$ *divides* $\beta$ *and* $0 \leq \beta < (n + r - 1) \cdot 2^i$.

*Proof.* By Proposition 5.1, $\beta$ can be expressed as a sum of coefficients all of which are multiples of $2^i$, and hence $\beta$ itself is a multiple of $2^i$. By Proposition 5.7, the only node whose interval contains negative values is the False node, and hence $\beta \geq 0$. Now, using Proposition 5.3, there must be an assignment to the variables $\{x_{0,1}, \ldots, x_{i,r-1}\}$ such that $2^0 \delta_{0,1} x_{0,1} + \cdots + 2^i \delta_{i,r-1} x_{i,r-1}$ belongs to the interval. Therefore:

$$\begin{aligned} \beta &\leq 2^0 \delta_{0,1} x_{0,1} + \cdots + 2^i \delta_{i,r-1} x_{i,r-1} \leq 2^0 + 2^0 + \cdots + 2^i \\ &= n2^0 + n2^1 + \cdots + n2^{i-1} + (r-1) \cdot 2^i = n(2^i - 1) + 2^i (r-1) \\ &< 2^i (n + r - 1) \end{aligned}$$

□

**Corollary 15.** *The number of nodes with selector variable* $x_{i,r}$ *is bounded by* $n + r - 1$. *In particular, the size of the ROBDD belongs to* $\mathcal{O}(n^2 m)$.

*Proof.* Let $\nu_1, \nu_2, \dots, \nu_t$ be all the nodes with selector variable $x_{i,r}$. Let $[\beta_j, \gamma_j]$ the interval of $\nu_j$. Note that such intervals are pair-wise disjoint since a non-empty intersection would imply that there exists a constraint represented by two different ROBDDs. Hence we can assume, without loss of generality, that $\beta_1 < \beta_2 < \cdots < \beta_t$. Due to Lemma 14, we know that $\beta_j - \beta_{j-1} \geq 2^i$. Hence $2^i(n+r-1) > \beta_t \geq \beta_{t-1} + 2^i \geq \cdots \geq \beta_1 + 2^i(t-1) \geq 2^i(t-1)$ and we can conclude that $t < n+r$. □

### 4.2 A Consistent Encoding for PB Constraints

Let us now take an arbitrary PB constraint $C : a_1x_1 + \cdots a_nx_n \leq K$ and assume that $a_M$ is the largest coefficient. For $m = \log a_M$, we can rewrite $C$ splitting the coefficients into powers of two as shown in Example 13:

$$\begin{array}{rccccccccl} \tilde{C}: & 2^0 \cdot \delta_{0,1} \cdot x_{0,1} & + & 2^0 \cdot \delta_{0,2} \cdot x_{0,2} & + & \cdots & + & 2^0 \cdot \delta_{0,n} \cdot x_{0,n} & + & \\ & 2^1 \cdot \delta_{1,1} \cdot x_{1,1} & + & 2^1 \cdot \delta_{1,2} \cdot x_{1,2} & + & \cdots & + & 2^1 \cdot \delta_{1,n} \cdot x_{1,n} & + & \\ & & & \cdots & & & & & + & \\ & 2^m \cdot \delta_{m,1} \cdot x_{m,1} & + & 2^m \cdot \delta_{m,2} \cdot x_{m,2} & + & \cdots & + & 2^m \cdot \delta_{m,n} \cdot x_{m,n} & \leq & K, \end{array}$$

where $\delta_{m,r}\ \delta_{m-1,r}\ \cdots\ \delta_{0,r}$ is the binary representation of $a_r$. Notice that $C$ and $\tilde{C}$ represent the same constraint if we add clauses expressing that $x_{i,r} = x_r$ for appropriate $i$ and $r$. This process is called *coefficient decomposition* of the PB constraint. A similar idea was given by Bartzis and Bultan (2003).

The important remark is that, using a consistent SAT encoding of the ROBDD for $\tilde{C}$ (e.g. the one given in Eén & Sörensson, 2006, or the one presented in Section 6) and adding clauses expressing that $x_{i,r} = x_r$ for appropriate $i$ and $r$, we obtain a consistent encoding for the original constraint $C$ using $\mathcal{O}(n^2 \log a_M)$ auxiliary variables and clauses.

This is not difficult to see. Take an assignment $A$ over the variables of $C$ which cannot be extended to a model of $C$. This is because the coefficients corresponding to the variables true in $A$ add more than $K$. Using the clauses for $x_{i,r} = x_r$, unit propagation will produce an assignment to the $x_{i,r}$'s that cannot be extended to a model of $\tilde{C}$. Since the encoding for $\tilde{C}$ is consistent, a false clause will be found. Conversely, if we consider an assignment $A$ over the variables of $C$ that can be extended to a model of $C$, this assignment can clearly be extended to a model for $\tilde{C}$ and the clauses expressing $x_{i,r} = x_r$. Hence, unit propagation on those clauses and the encoding of $\tilde{C}$ will not detect a false clause.

**Example 16.** *Consider the PB constraint $C : 2x_1 + 3x_2 + 5x_3 \leq 6$. For obtaining the consistent encoding we have presented, we first rewrite $C$ by splitting the coefficients into powers of two:*

$$C' : 1x_{0,2} + 1x_{0,3} + 2x_{1,1} + 2x_{1,2} + 4x_{2,3} \leq 6.$$

*Next, we encode $C'$ into a ROBDD and finally encode the ROBDD into SAT and add clauses for enforcing the relations $x_{i,j} = x_j$. Or, instead of that, we can replace $x_{i,j}$ by $x_j$ into the ROBDD, and encode the result into SAT. Figure 6 shows the decision diagram after the replacement.*

### 4.3 A Generalized Arc-consistent Encoding for PB Constraints

Unfortunately, the previous approach does not result in a GAC encoding. The intuitive idea can be seen in the following example:

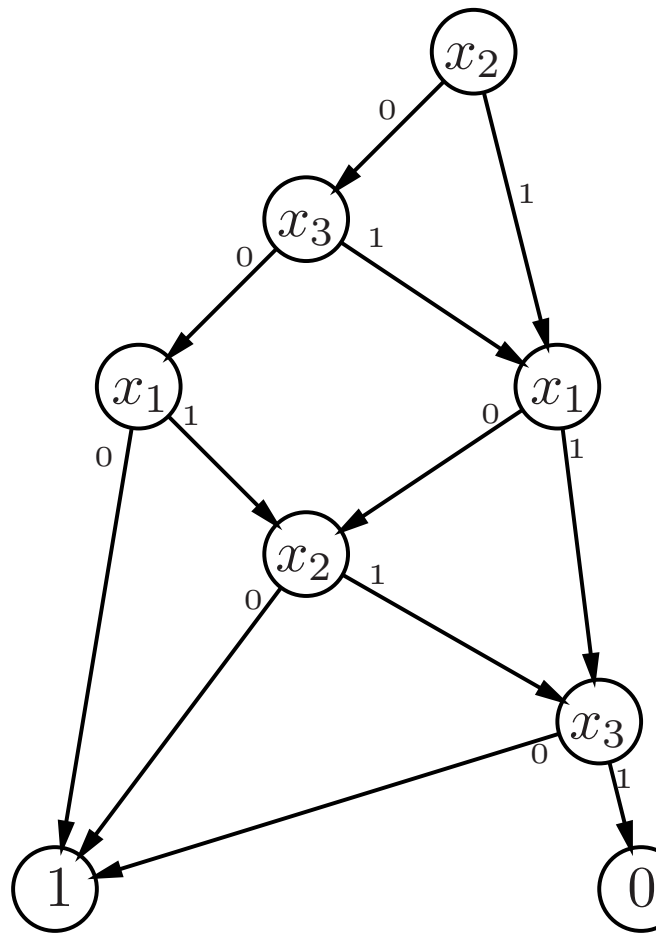

Figure 6: Decision Diagram of $2x_1 + 3x_2 + 5x_3 \leq 6$ after decomposing the coefficients into powers of two.

**Example 17.** *Let us consider the constraint $3x_1 + 4x_2 \leq 6$. After splitting the coefficients into powers of two, we obtain $C' : x_{0,1} + 2x_{1,1} + 4x_{2,2} \leq 6$. If we set $x_{2,2}$ to true, $C'$ implies that either $x_{0,1}$ or $x_{1,1}$ have to be false, but the encoding cannot exploit the fact that both variables will receive the same truth value and hence both should be propagated. Adding clauses stating that $x_{0,1} = x_{1,1}$ does not help in this sense.*

In order to overcome this limitation, we follow the method presented by Bessiere, Katsirelos, Narodytska, and Walsh (2009) and Bailleux et al. (2009). Let $C : a_1x_1 + \cdots + a_nx_n \leq K$ be an arbitrary PB constraint. We denote as $C_i$ the constraint $a_1x_1 + \cdots + a_i \cdot 1 + \cdots + a_nx_n \leq K$, i.e., the constraint assuming $x_i$ to be true. For every $i$ with $1 \leq i \leq n$, we encode $C_i$ as in Section 4.2 and, in addition, we add the binary clause $r_i \vee \overline{x}_i$, where $r_i$ is the root of the ROBDD for $C_i$. This clause helps us to preserve GAC: given an assignment $A$ such that $A \cup \{x_i\}$ cannot be extended to a model of $C$, literal $\overline{r}_i$ will be propagated using $A$ (because the encoding for $C_i$ is consistent). Hence the added clause will allow us to propagate $\overline{x}_i$.

**Example 18.** *Consider again the PB constraint $C : 2x_1 + 3x_2 + 5x_3 \leq 6$. Let us define the constraints $C_1 : 3x_2 + 5x_3 \leq 4$, $C_2 : 2x_1 + 5x_3 \leq 3$ and $C_3 : 2x_1 + 3x_2 \leq 1$. Now, we encode these constraints into ROBDDs as in the previous section, with coefficient decomposition. Figure 7 shows the resulting ROBDDs. Finally, we need to encode them into SAT consistently, and then add the clauses $r_i \vee \overline{x}_i$, assuming that the variable associated with the root of the ROBDD for $C_i$ is $r_i$.*

*This encoding is GAC: take for instance the assignment $A = \{x_1 = 1\}$. Constraint $C_3$ is not satisfied. Hence, by consistency, $\overline{r}_3$ is propagated. Therefore, clause $r_3 \vee \overline{x}_3$ propagates $\overline{x}_3$, as wanted. The propagation with other assignments is similar.*

All in all, the suggested encoding is GAC and uses $\mathcal{O}(n^3 \log(a_M))$ clauses and auxiliary variables, where $a_M$ is the largest coefficient.

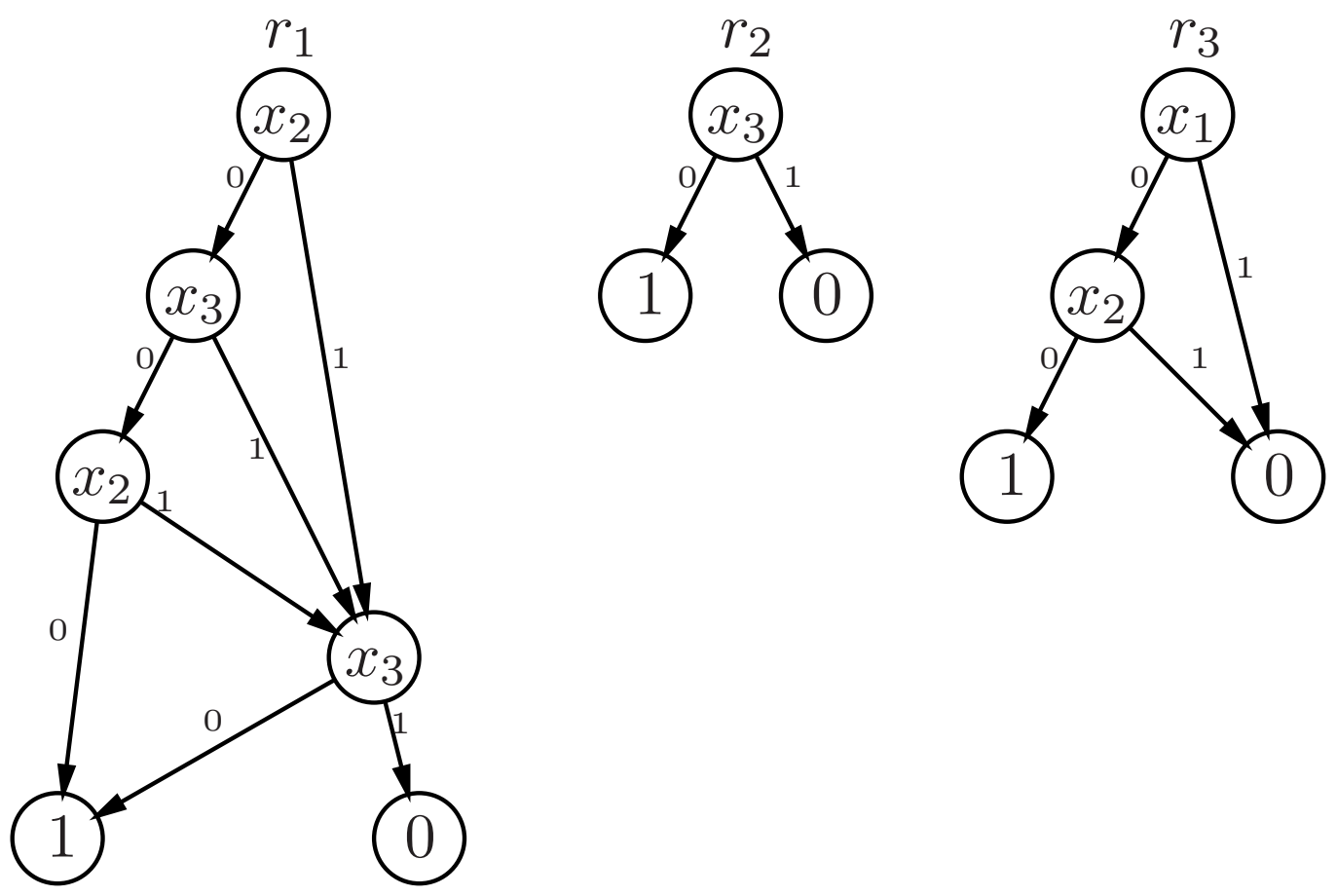

Figure 7: ROBDDs[2] of $C_1$, $C_2$ and $C_3$ with coefficient decomposition.

## 5. An Algorithm for Constructing ROBDDs for Pseudo-Boolean Constraints

Let us fix a Pseudo-Boolean constraint $a_1x_1 + \cdots + a_nx_n \leq K$ and a variable ordering $[x_1, x_2, \ldots, x_n]$. The goal of this section is to prove that one can construct the ROBDD of this constraint using this ordering in polynomial time with respect to the ROBDD size and $n$.

This algorithm builds standard ROBDDs, but it can be used to build ROBDDs with coefficient decomposition: we just need to first split the coefficients and, secondly, apply the algorithm. Forthcoming Example 21 shows in detail the overall process. A very similar version of this algorithm was described by Mayer-Eichberger (2008).

The key point of the algorithm will be to label each node of the ROBDD with its interval. In the following, for every $i \in \{1, 2, \ldots, n+1\}$, we will use a set $L_i$ consisting of pairs $([\beta, \gamma], \mathcal{B})$, where $\mathcal{B}$ is the ROBDD of the constraint $a_ix_i + \cdots + a_nx_n \leq K'$ for every $K' \in [\beta, \gamma]$ (i.e., $[\beta, \gamma]$ is the interval of $\mathcal{B}$). All these sets will be kept in a tuple $\mathcal{L} = (L_1, L_2, \ldots, L_{n+1})$.

Note that by definition of the ROBDD's intervals, if $([\beta_1, \gamma_1], \mathcal{B}_1)$ and $([\beta_2, \gamma_2], \mathcal{B}_2)$ belong to $L_i$ then either $[\beta_1, \gamma_1] = [\beta_2, \gamma_2]$ or $[\beta_1, \gamma_1] \cap [\beta_2, \gamma_2] = \emptyset$. Moreover, the first case holds if and only if $\mathcal{B}_1 = \mathcal{B}_2$. Therefore, $L_i$ can be represented with a *binary search tree-like* data structure, where insertions and searches can be done in logarithmic time. The function **search**$(K, L_i)$ searches whether there exists a pair $([\beta, \gamma], \mathcal{B}) \in L_i$ with $K \in [\beta, \gamma]$. Such a tuple is returned if it exists, otherwise an empty interval is returned in the first component of the pair. Similarly, we will also use function **insert**$(([\beta, \gamma], \mathcal{B}), L_i)$ for insertions. The overall algorithm is detailed in Algorithm 1 and Algorithm 2:

**Theorem 19.** *Algorithm 1 runs in $\mathcal{O}(nm \log m)$ time (where $m$ is the size of the ROBDD) and is correct in the following sense:*

2. Actually, the diagram after replacing the variables $x_{i,j}$ by $x_j$ is not a ROBDD. However, we will denote them as ROBDDs for simplicity.

**Algorithm 1** Construction of ROBDD algorithm

**Require:** Constraint $C : a_1x_1 + \ldots + a_nx_n \leq K'$
**Ensure:** returns $\mathcal{B}$ the ROBDD of $C$.

**for all** $i$ such that $1 \leq i \leq n+1$ **do**
  $L_i \leftarrow \Big\{ \big((-\infty, -1], False\big),\ \big([a_i + a_{i+1} + \cdots + a_n, \infty), True\big) \Big\}$
**end for**
$\mathcal{L} \leftarrow (L_1, \ldots, L_{n+1})$.
$([\beta, \gamma], \mathcal{B}) \leftarrow$ **BDDConstruction**$(1, a_1x_1 + \ldots + a_nx_n \leq K', \mathcal{L})$.
**return** $\mathcal{B}$.

**Algorithm 2** Procedure BDDConstruction

**Require:** integer $i \in \{1, 2, \ldots, n+1\}$, constraint $C : a_ix_i + \ldots + a_nx_n \leq K'$ and set $\mathcal{L}$
**Ensure:** returns $[\beta, \gamma]$ interval of $C$ and $\mathcal{B}$ its ROBDD

$([\beta, \gamma], \mathcal{B}) \leftarrow$ **search**$(K', L_i)$.
**if** $[\beta, \gamma] \neq \emptyset$ **then**
  **return** $([\beta, \gamma], \mathcal{B})$
**else**
  $([\beta_F, \gamma_F], \mathcal{B}_F) :=$ **BDDConstruction**$(i+1, a_{i+1}x_{i+1} + \ldots + a_nx_n \leq K', \mathcal{L})$.
  $([\beta_T, \gamma_T], \mathcal{B}_T) :=$ **BDDConstruction**$(i+1, a_{i+1}x_{i+1} + \ldots + a_nx_n \leq K' - a_i, \mathcal{L})$.
  **if** $[\beta_T, \gamma_T] = [\beta_F, \gamma_F]$ **then**
    $\mathcal{B} \leftarrow \mathcal{B}_T$
    $[\beta, \gamma] \leftarrow [\beta_T + a_i, \gamma_T]$
  **else**
    $\mathcal{B} \leftarrow$ **ite**$(x_i, \mathcal{B}_T, \mathcal{B}_F)$ // root $x_i$, true branch $\mathcal{B}_T$ and false branch $\mathcal{B}_F$.
    $[\beta, \gamma] \leftarrow [\beta_F, \gamma_F] \cap [\beta_T + a_i, \gamma_T + a_i]$
  **end if**
  **insert**$(([\beta, \gamma], \mathcal{B}), L_i)$
  **return** $([\beta, \gamma], \mathcal{B})$
**end if**

1. *$K'$ belongs to the interval returned by* **BDDConstruction**$(a_ix_i+\cdots+a_nx_n \leq K', \mathcal{L})$.

2. *The tuple $([\beta,\gamma],\mathcal{B})$ returned by* **BDDConstruction** *consist of a BDD $\mathcal{B}$ and its interval $[\beta,\gamma]$.*

3. *If* **BDDConstruction** *returns $([\beta,\gamma],\mathcal{B})$, then the BDD $\mathcal{B}$ is reduced.*

*Proof.* Let us first start with the three correctness statements:

1. If $K'$ is found in $L_i$ at line 1 of Algorithm 2 the statement is obviously true. Otherwise let us reason by induction on $i$. The base case is when $i = n+1$, and since $L_{n+1}$ contains the intervals $(-\infty,-1]$ and $[0,\infty]$, the **search** call at line 1 will succeed and hence the result holds. For $i < n+1$ we can assume, by induction hypothesis, that $K' \in [\beta_F,\gamma_F]$ and $K'-a_i \in [\beta_T,\gamma_T]$. If the two intervals coincide the result is obvious, otherwise it is also easy to see that $K' \in [\beta_F,\gamma_F] \cap [\beta_T+a_i,\gamma_T+a_i]$.

2. Let us prove that in every moment all the tuples of $\mathcal{L}$ are correct, i.e., they contain BDDs with their correct interval. Since the returned value is always an element of some $L_i$, this proves the statement.

   By Proposition 5.6 and 5.7, initial tuples of $\mathcal{L}$ are correct. We have to prove that if all the previously inserted intervals are correct, the current interval is also correct. It follows in virtue of Proposition 7.

3. Let us prove that all the tuples of $\mathcal{L}$ contain only reduced BDDs. As before, all the initial BDDs in $\mathcal{L}$ are reduced. Let $\mathcal{B}$ be a BDD computed by the algorithm, with children $\mathcal{B}_T$ and $\mathcal{B}_F$. By induction hypothesis, they are reduced, so $\mathcal{B}$ is reduced if and only if its two children are not equal. The algorithm creates a node only if its children's intervals are different. Therefore, $\mathcal{B}_T$ and $\mathcal{B}_F$ do not represent the same Boolean constraint, so they are different BDDs.

Regarding runtime, since the cost of search and insertion in $L_i$ is logarithmic in its size, the cost of the algorithm is $\mathcal{O}(\log m)$ times the number of calls to **BDDConstruction**. Hence, it only remains to show that there are at most $\mathcal{O}(nm)$ calls.

Every call (but the first one) to **BDDConstruction** is done when we are exploring an edge of the ROBDD. Notice that no edge is explored twice, since the edges are only explored from the parent node and whenever we reach an explored node there are no recursive calls to **BDDConstruction**. On the other hand, for every edge of the ROBDD we make $2k-1$ calls, where $k$ is the *length* of the edge (if the nodes joined by the edge have variables $x_i$ and $x_j$ we say that its *length* is $|i-j|$). Since the ROBDD has $\mathcal{O}(m)$ edges and their length is $\mathcal{O}(n)$, the number of calls is $\mathcal{O}(nm)$. □

Let us illustrate the algorithm with an example:

**Example 20.** *Take the constraint $C : 2x_1+3x_2+5x_3 \leq 6$, and let us apply the algorithm to obtain the ROBDD in the ordering $[x_1,x_2,x_3]$. Figure 8 represents the recursive calls to BDDConstruction and the returned parameters (the ROBDD and the interval).*

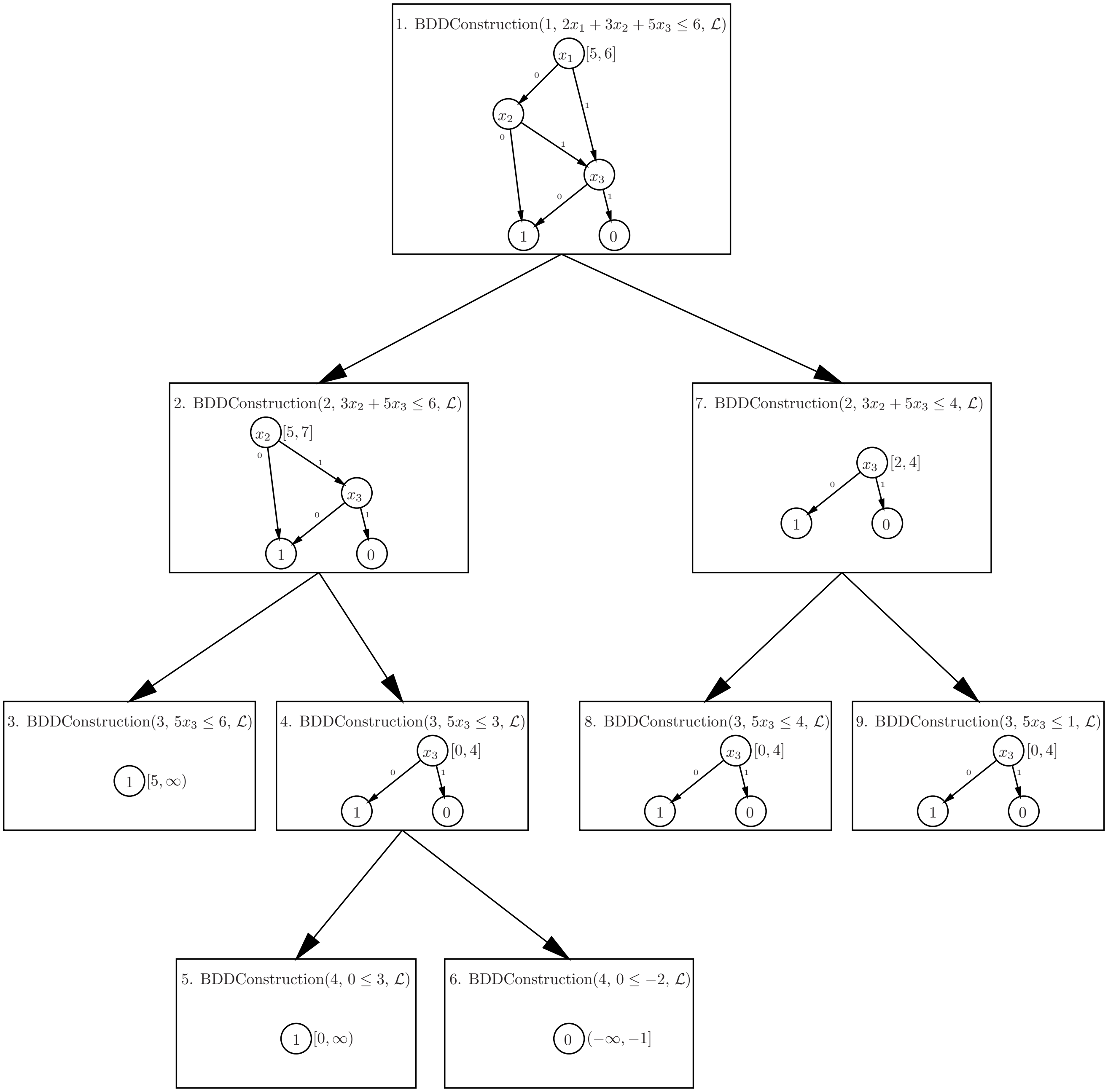

Figure 8: Recursive calls to **BDDConstruction**, with the returned values.

- *In calls number 3, 5, 6, 8 and 9, the search function returns true and the interval and the ROBDD are returned without any other computation.*
- *In call number 7, the two recursive calls return the same interval (and, therefore, the same ROBDD). Hence, that ROBDD is returned.*
- *In call number 1 the two recursive calls return two different ROBDDs, so it adds a node to join the two ROBDDs into another one, which is returned. The same happens in calls number 2 and 4.*

The overall process with coefficient decomposition would work as in the following example:

**Example 21.** *Let us take the constraint* $C : 2x_1 + 3x_2 + 5x_3 \leq 6$*. If we want to build the ROBDD with coefficient decomposition using Algorithm 1, we proceed as follows:*

1. *Split the coefficients and obtain* $C' : 1y_1 + 1y_2 + 2y_3 + 2y_4 + 4y_5 \leq 6$*, where* $x_1 = y_3$*,* $x_2 = y_1 = y_4$ *and* $x_3 = y_2 = y_5$*.*
2. *Apply the algorithm to* $C'$ *and obtain a ROBDD* $\mathcal{B}'$*.*
3. *Replace* $y_1$ *for* $x_2$*,* $y_2$ *for* $x_3$*, etc. in the nodes of* $\mathcal{B}'$*.*

## 6. SAT Encodings of BDDs for Monotonic Functions

In this section we consider a BDD representing a monotonic function $F$ and we want to encode it into SAT. As expected, we want the encoding to be as small as possible and GAC.

The encoding explained here is valid with any type of BDDs, so, in particular, it is valid with ROBDDs. The main differences with the Minisat+ encoding (Eén & Sörensson, 2006) is the number of clauses generated (6 ternary clauses per node versus one binary and one ternary clauses per node) and that our encoding is GAC with any variable ordering.

As usual, the encoding introduces an auxiliary variable for every node. Let $\nu$ be a node with selector variable $x$ and auxiliary variable $n$. Let $f$ be the variable of its false child and $t$ be the variable of its true child. Only two clauses per node are needed:

$$\overline{f} \rightarrow \overline{n} \qquad \overline{t} \wedge x \rightarrow \overline{n}.$$

Furthermore, we add a unit clause with the variable of the True node and another one with the negation of the variable of the False node.

**Theorem 22.** *The encoding is consistent in the following sense: a partial assignment* $A$ *cannot be extended to a model of* $F$ *if and only if* $\overline{r}$ *is propagated by unit propagation, where* $r$ *is the root of the BDD.*

*Proof.* We prove the theorem by induction on the number of variables of the BDD. If the BDD has no variables, then the BDD is either the True node or the False node and the result is trivial.

Assume that the result is true for BDDs with less than $k$ variables, and let $F$ be a function whose BDD has $k$ variables. Let $r$ be the root node, $x_1$ its selector variable and $f, t$ respectively its false and true children (note that we abuse the notation and identify nodes with their auxiliary variable). We denote by $F_1$ the function $F_{|x_1=1}$ (i.e., $F$ after setting $x_1$ to true) and by $F_0$ the function $F_{|x_1=0}$.

- Let $A$ be a partial assignment that cannot be extended to a model of $F$.
  - Assume $x_1 \in A$. Since $A$ cannot be extended, the assignment $A \setminus \{x_1\}$ cannot be extended to a model of $F_1$. By definition of the BDD, the function $F_1$ has $t$ as a BDD. By induction hypothesis, $\bar{t}$ is propagated, and since $x_1 \in A$, $\bar{r}$ is also propagated.
  - Assume $x_1 \notin A$. Then, the assignment $A \setminus \{\overline{x}_1\}$ cannot be extended to a model of $F_0$. Since $F_0$ has $f$ as a BDD, by induction hypothesis $\overline{f}$ is propagated, and hence $\bar{r}$ also is.
- Let $A$ be a partial assignment, and assume $\bar{r}$ has been propagated. Then, either $\overline{f}$ has also been propagated or $\bar{t}$ has been propagated and $x_1 \in A$ (note that $x_1$ has not been propagated because it only appears in one clause which is already true).
  - Assume that $\overline{f}$ has been propagated. Since $f$ is the BDD of $F_0$, by induction hypothesis the assignment $A \setminus \{x_1, \overline{x}_1\}$ cannot be extended to a model of $F_0$. Since the function is monotonic, neither can $A \setminus \{x_1, \overline{x}_1\}$ be extended to a model of $F$. Therefore, $A$ cannot be extended to a model of $F$.
  - Assume that $\bar{t}$ has been propagated and $x_1 \in A$. Since $t$ is the BDD of $F_1$, by induction hypothesis $A \setminus \{x_1\}$ cannot be extended to a model of $F_1$, so neither can $A$ be extended to a model of $F$.

□

For obtaining a GAC encoding, we only have to add a unit clause.

**Theorem 23.** *If we add a unit clause forcing the variable of the root node to be true, the previous encoding becomes generalized arc-consistent.*

*Proof.* We will prove it by induction on the variables of the BDD. The case with zero variables is trivial, so let us prove the induction case.

As before, let $r$ be the root node, with $x_1$ its selector variable and $n$ its auxiliary variable, and $f, t$ its false and true children. We denote by $F_0$ and $F_1$ the functions $F_{|x_1=0}$ and $F_{|x_1=1}$.

Let $A$ be a partial assignment that can be extended to a model of $F$. Assume that $A \cup \{x_i\}$ cannot be extended. We want to prove that $\overline{x}_i$ will be propagated.

- Let us assume that $x_1 \in A$. In this case, $t$ is propagated due to the clause $\bar{t} \wedge x_1 \rightarrow \bar{n}$ and the unit clause $n$. Since $x_1 \in A$ and $A \cup \{x_i\}$ cannot be extended to a model of $F$, $A \setminus \{x_1\} \cup \{x_i\}$ neither can be extended to an assignment satisfying $F_1$. By induction hypothesis, since $t$ is the BDD of the function $F_1$, $\overline{x}_i$ is propagated.
- Let us assume that $x_1 \notin A$ and $x_i \neq x_1$. Since $F$ is monotonic, $A \cup \{x_i\}$ cannot be extended to a model of $F$ if and only if it cannot be extended to a model of $F_0$. Notice that $f$ is propagated thanks to the clause $\overline{f} \rightarrow \bar{n}$ and the unit clause $n$. By induction hypothesis, the method is GAC for $F_0$, so $\overline{x}_i$ is propagated.

- Finally, assume that $x_1 \notin A$ and $x_i = x_1$. Since $A \cup \{x_1\}$ cannot be extended to a model of $F$, $A$ cannot be extended to a model of $F_1$. By Theorem 22, $\bar{t}$ is propagated and, due to $\bar{t} \wedge x_1 \rightarrow \overline{n}$ and $n$, also is $\overline{x}_1$.

□

We finish this section with an example illustrating how the suggested encoding of BDDs into SAT can be used in the different PB encoding methods we have presented in this paper.

**Example 24.** *Consider the constraint $C : 2x_1 + 3x_2 + 5x_3 \leq 6$. We will encode this constraint into SAT with three methods: with the usual ROBDD encoding; with the consistent approach of ROBDDs and splitting of the coefficients, explained in Section 4.2; and with the GAC approach of ROBDDs and splitting of the coefficients explained in Section 4.3.*

1. *BDD-1: this method builds the ROBDD for C and then encodes it into SAT. Hence we start by building the ROBDD of C, which can be seen in the last picture of Figure 1. Now, we need to encode it into SAT. Let $y_1$, $y_2$ and $y_3$ be fresh variables corresponding to the nodes of the ROBDD of C having respectively $x_1$, $x_2$ and $x_3$ as selector variable.*

   *For node $y_1$, we have to add the clauses $\overline{y_2} \rightarrow \overline{y_1}$ and $x_1 \wedge \overline{y_3} \rightarrow \overline{y_1}$.*

   *For $y_2$, we have to add the clauses $\overline{\top} \rightarrow \overline{y_2}$ and $x_2 \wedge \overline{y_3} \rightarrow \overline{y_2}$, where $\top$ is the tautology symbol.*

   *For $y_3$, we have to add the clauses $\overline{\top} \rightarrow \overline{y_3}$ and $x_3 \wedge \overline{\bot} \rightarrow \overline{y_3}$, where $\bot$ is the contradiction symbol.*

   *Moreover, we have to add the unit clauses $\top$, $\overline{\bot}$ and $y_1$. All in all, after removing the units and tautologies, the clauses obtained are $y_1$, $y_2$, $\overline{x_1} \vee y_3$, $\overline{x_2} \vee y_3$ and $\overline{x_3} \vee \overline{y_3}$.*

2. *BDD-2: we build the ROBDD of C with coefficient decomposition as in Example 21. Figure 6 shows the resulting ROBDD. We introduce variables $y_1, y_2, \ldots, y_6$ for every node of the ROBDD. More precisely, the first $x_2$ node (starting top-down) receives variable $y_1$, the next $x_2$ node gets $y_5$. The first $x_3$ receives $y_2$ and the other one $y_6$. Finally the leftmost $x_1$ node gets variable $y_3$ and the other one $y_4$. We have to add the following clauses: $\overline{y_2} \rightarrow \overline{y_1}$, $\overline{y_4} \wedge x_2 \rightarrow \overline{y_1}$, $\overline{y_3} \rightarrow \overline{y_2}$, $\overline{y_4} \wedge x_3 \rightarrow \overline{y_2}$, $\overline{\top} \rightarrow \overline{y_3}$, $\overline{y_5} \wedge x_1 \rightarrow \overline{y_3}$, $\overline{y_5} \rightarrow \overline{y_4}$, $\overline{y_6} \wedge x_1 \rightarrow \overline{y_4}$, $\overline{\top} \rightarrow \overline{y_5}$, $\overline{y_6} \wedge x_2 \rightarrow \overline{y_5}$, $\overline{\top} \rightarrow \overline{y_6}$, $\overline{\bot} \wedge x_3 \rightarrow \overline{y_6}$, and the unit clauses $\top$, $\overline{\bot}$ and $y_1$.*

   *After removing the units from the clauses and tautologies, we obtain $y_1$, $y_2$, $y_3$, $y_4 \vee \overline{x_2}$, $y_4 \vee \overline{x_3}$, $y_5 \vee \overline{x_1}$, $y_5 \vee \overline{y_4}$, $y_6 \vee \overline{x_1} \vee \overline{y_4}$, $y_6 \vee \overline{x_2} \vee \overline{y_5}$ and $\overline{x_3} \vee \overline{y_6}$.*

   *Notice that this encoding is consistent: if we have the assignment $A = \{x_2, x_3\}$, then $y_4$ is propagated by the clause $y_4 \vee \overline{x_2}$, which in turn propagates $y_5$ due to clause $y_5 \vee \overline{y_4}$ and finally $y_6$ is propagated by the clause $y_6 \vee \overline{x_2} \vee \overline{y_5}$. A contradiction is found with clause $\overline{x_3} \vee \overline{y_6}$.*

   *However, the encoding is not GAC: the partial assignment $A = \{x_1\}$ can only propagate $y_5$. However, $\overline{x_3}$ should also be propagated.*

3. *BDD-3: let $C_1$, $C_2$ and $C_3$ be the constraints setting respectively $x_1$, $x_2$ and $x_3$ to true. Figure 7 shows the ROBDDs of these constraints. We have to encode these ROBDDs* *as usual, as in BDD-2, but replacing the unit clause $r$ of the root by $\overline{r} \rightarrow \overline{x_i}$. In this case the variables associated with the roots of $C_1, C_2$ and $C_3$ will be $y_1$, $z_1$ and $w_1$ respectively.*

*After removing the units and tautologies, clauses from $C_1$ are $y_1 \vee \overline{x_1}$, $y_2 \vee \overline{y_1}$, $y_4 \vee \overline{x_2} \vee \overline{y_1}$, $y_3 \vee \overline{y_2}$, $y_4 \vee \overline{x_3} \vee \overline{y_2}$, $y_4 \vee \overline{x_2} \vee \overline{y_3}$ and $\overline{x_3} \vee \overline{y_4}$.*

*Clauses from $C_2$ are $z_1 \vee \overline{x_2}$ and $\overline{x_3} \vee \overline{z_1}$.*

*Finally, clauses from $C_3$ are $w_1 \vee \overline{x_3}$, $w_2 \vee \overline{w_1}$, $\overline{x_1} \vee \overline{w_1}$ and $\overline{x_2} \vee \overline{w_2}$.*

*This encoding is GAC. Take, for instance, the assignment $A = \{x_1\}$. In this case, $\overline{w_1}$ is propagated in virtue of $\overline{x_1} \vee \overline{w_1}$ and $\overline{x_3}$ is propagated by clause $w_1 \vee \overline{x_3}$.*

## 7. Related Work

Due to the ubiquity of Pseudo-Boolean constraints and the success of SAT solvers, the problem of encoding those constraints into SAT has been thoroughly studied in the literature. In the following we review the most important contributions, paying special attention to the basic idea on which they are based, the encoding size, and the propagation properties the encodings fulfill. To ease the presentation, in the remaining of this section we will always assume that the constraint we want to encode is $a_1x_1 + \ldots + a_nx_n \leq k$, with maximum coefficient $a_{max}$.

The first encoding to mention is the one proposed by Warners (1998). In a nutshell, the encoding uses several adders for numbers in binary representation. First of all, the left hand side of the constraint is split into two halves, each of which is recursively treated to compute the corresponding partial sum. After that, the two partial sums are added and the final result is compared with $k$ . The encoding uses $O(n \log(a_{max}))$ clauses and variables and is neither consistent nor GAC. This is not surprising, since adders for numbers in binary make extensive use of xors, which do not have good propagation properties.

Bailleux et al. (2006) introduced an encoding "very close to those using a BDD and translating it into clauses". In order to understand the differences between their construction and BDDs let us introduce it in detail. First of all, the coefficients are ordered from small to large. Then, the root is labeled with variable $D_{n,k}$, expressing that the sum of the first $n$ terms is no more than $k$. Its two children are $D_{n-1,k}$ and $D_{n-1,k-a_n}$, which correspond to setting $x_n$ to false and true, respectively. The process is repeated until nodes corresponding to trivial constraints are reached, which are encoded as true or false. For each node $D_{i,b}$ with children $D_{i-1,b}$ and $D_{i-1,b-a_i}$, the following four clauses are added:

$$\begin{array}{ll} D_{i-1,b-a_i} \rightarrow D_{i,b} & \overline{D}_{i-1,b} \rightarrow \overline{D}_{i,b} \\ \overline{D}_{i-1,b-a_i} \wedge x_i \rightarrow \overline{D}_{i,b} & D_{i-1,b} \wedge \overline{x}_i \rightarrow D_{i,b} \end{array}$$

**Example 25.** *The encoding proposed by Bailleux et al. (2006) on $2x_1 + \cdots + 2x_{10} + 5x_{11} + 6x_{12} \leq 10$ is illustrated in Figure 9. Node $D_{10,5}$ represents $2x_1 + 2x_2 + \cdots + 2x_{10} \leq 5$, whereas node $D_{10,4}$ represents $2x_1 + 2x_2 + \cdots 2x_{10} \leq 4$. The method fails to identify that these two PB constraints are equivalent and hence subtrees B and C will not be merged, yielding a much larger representation than with ROBDDs.*

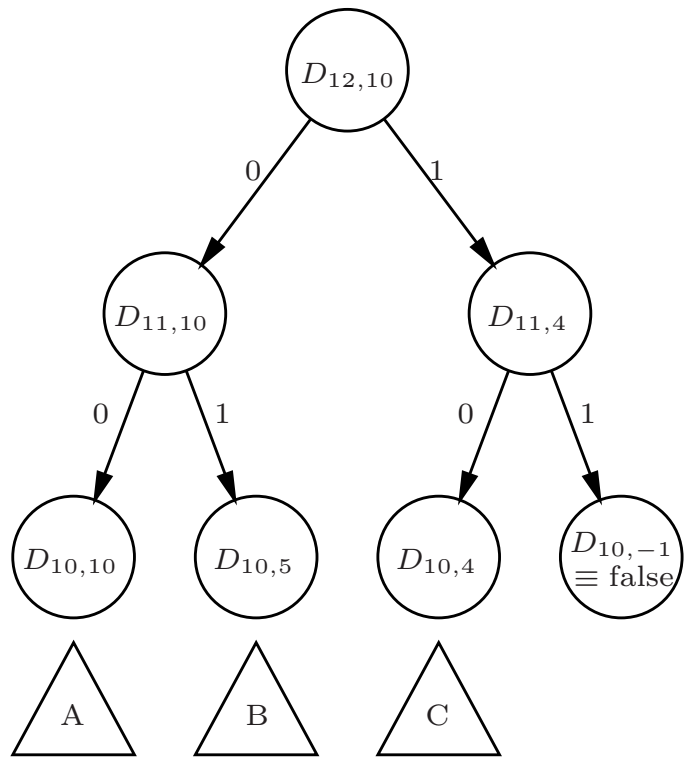

Figure 9: Tree-like construction of Bailleux et al. (2006) for $2x_1+\cdots+2x_{10}+5x_{11}+6x_{12}\leq 10$

The resulting encoding is GAC, but an example of a PB constraint family is given for which their kind of *non-reduced* BDDs, with *their concrete variable ordering* is exponentially large. However, as we have shown in Section 3.2, ROBDDs for this family are polynomial.

Several important new contributions were presented in the paper by the MiniSAT team (Eén & Sörensson, 2006). The paper describes three encodings, all of which are implemented in the MiniSAT+ tool. The first one is a standard ROBDD construction for Pseudo-Boolean constraints. This is done in two steps: first, they suggest a simple dynamic programming algorithm for constructing a non-reduced BDD, which is later reduced. The result is a ROBDD, but the first step may take exponential time even if the final ROBDD is polynomial. Once the ROBDD is constructed, they suggest to encode it into SAT using 6 ternary clauses per node. The paper showed that, given a concrete variable ordering, the encoding is GAC. Regarding the ROBDD size, the authors cite the work of Bailleux et al. (2006) to state the BDDs are exponential in the worst case. As we have seen before, the citation is not correct because Bailleux et al do not construct ROBDDs.

The second method is similar to the one of Warners (1998) in the sense that the construction relies on a network of adders. First of all coefficients are decomposed into binary representation. For each bit $i$, a bucket is created with all variables whose coefficient has bit $i$ set to one. The $i$-th bit of the left-hand side of the constraint is computed using a series of full adders and half adders. Finally, the resulting sum is lexicographically compared to $k$. The resulting encoding is neither consistent nor GAC and uses a number of adders linear in the sum of the number of digits of the coefficients.

The last method they suggest is the use of sorting networks. Numbers are expressed in unary representation and coefficients are decomposed using a mixed radix representation. The smaller the number in this representation, the smaller the encoding. In this setting, sorting networks are used to play the same role of adders, but with better propagation properties. If $N$ is smaller than the sum of the digits of all coefficients in base 2, the size of the encoding is $O(N \log^2 N)$. Whereas this encoding is not GAC for arbitrary Pseudo-Boolean constraints, generalized arc-consistency is obtained for cardinality constraints.

| Encoding | Reference | Clauses | Consist. | GAC |
|---|---|---|---|---|
| Warners | (Warners, 1998) | $\mathcal{O}(n \log a_{max})$ | NO | NO |
| Non-reduced BDD | (Bailleux et al., 2006) | Exponential | YES | YES |
| ROBDD | (Eén & Sörensson, 2006) | Exponential (6 per node) | YES | YES |
| Adders | (Eén & Sörensson, 2006) | $\mathcal{O}(\sum \log a_i)$ | NO | NO |
| Sorting Networks | (Eén & Sörensson, 2006) | $\mathcal{O}((\sum \log a_i) \log^2(\sum \log a_i)$ | YES | NO |
| Watch Dog (WD) | (Bailleux et al., 2009) | $\mathcal{O}(n^2 \log n \log a_{max})$ | YES | NO |
| Gen. Arc-cons. WD | (Bailleux et al., 2009) | $\mathcal{O}(n^3 \log n \log a_{max})$ | YES | YES |

Table 1: Summary comparing the different encodings.

The first polynomial and GAC encoding for Pseudo-Boolean constraints, called Watch-Dog, was introduced by Bailleux et al. (2009). It uses $O(n^2 \log n \log a_{max})$ variables and $O(n^3 \log n \log a_{max})$ clauses. Again, numbers are expressed in unary representation and totalizers are used to play the role of sorting networks. In order to make the comparison with the right hand side trivial, the left-hand side and $k$ are incremented until $k$ becomes a power of two. Then, all coefficients are decomposed in binary representation and each bit is added independently, taking into account the corresponding carry. In the same paper, another encoding which is only consistent and uses $O(n \log n \log a_{max})$ variables and $O(n^2 \log n \log a_{max})$ clauses is also presented.

Finally, it is worth mentioning the work of Bartzis and Bultan (2003). The authors deal with the more general case in which the variables $x_i$ are not Boolean, but bounded integers $0 \leq x_i < 2^b$. They suggest a BDD-based approach very similar in flavor to our method of Section 4, but instead of decomposing the coefficients as we do, they decompose the variables $x_i$ in binary representation. The BDD ordering starts with the first bit of $x_1$, then the first bit of the $x_2$, etc... After that, the second bit is treated in a similar fashion, and so on. The resulting BDD has $O(n \cdot b \cdot \sum a_i)$ nodes and nothing is mentioned about propagation properties. For the case of Pseudo-Boolean constraints, i.e. $b = 1$, their approach amounts to standard BDDs.

Table 1 summarizes the different encodings of PB constraints into SAT.

## 8. Experimental Results

The goal of this section is to assess the practical interest of the encodings we have presented in the paper. Our aim is to evaluate to which extent BDD-based encodings are interesting from the practical point of view. For us, this means to study whether they are competitive with existing techniques, whether they show good behavior in general or are only interesting for very specific types of problems, or whether they produce smaller encodings.

For that purpose, first of all, we compare our encodings with other SAT encodings in terms of encoding time, number of clauses and number of variables. After that, we also consider total runtime (that is, encoding time plus solving time) of these encodings and we compare it with the runtime of state-of-the-art Pseudo-Boolean solvers. Finally, we briefly report on some experiments with sharing, that is, trying to encode several Pseudo-Boolean constraints in a single ROBDD.

All experiments were performed on a 2Ghz Linux Quad-Core AMD with a time limit of 1800 seconds per benchmark. The benchmarks used for these experiments were obtained from the Pseudo-Boolean Competition 2011 (`http://www.cril.univ-artois.fr/PB11/`), category *no optimization, small integers, linear constraints (DEC-SMALLINT-LIN)*. For compactness and clarity, we have grouped benchmarks that come from the same source into families. However, individual results can be found at `http://www.lsi.upc.edu/~iabio/BDDs/results.ods`.

### 8.1 The Bergmann-Hommel Test

In order to summarize the experiments and make it easier to extract conclusions, every experiment is accompanied with a *Bergmann-Hommel non-parametric hypothesis test* (Bergmann & Hommel, 1988) of the results with a confidence level of 0.1.

The Bergmann-Hommel test is a way of comparing the results of $n$ different methods over multiple independent data sets. It gives us two interesting pieces of information. First of all, it sorts the methods by giving them a real number between 1 and $n$, such that the lower the number the better the method. Moreover, it indicates, for each pair of methods, whether one method significantly improves upon the other. As an example, Figure 10 is the output of a Bergmann-Hommel test. BDD-1 is the best method but there is not significant difference between this method and BDD-2 (this is illustrated by a thick line connecting the methods). On the other hand, the Bergmann-Hommel test indicates that BDD-1 is significantly better than Adder, since there is no thick line connecting BDD-1 and Adder. The same can be said for BDD-1 and WD-1, BDD-1 and BDD-3, BDD-1 and WD-2, BDD-2 and Adder, etc.

We will now give a quick overview of how a Bergmann-Hommel test is computed. The remaining of this section can be skipped if the reader is not interested in the details of the test. On the other hand, for more detailed information, we refer the reader to the work of Bergmann and Hommel (1988).

Let us assume we have $n$ methods and $m$ data sets, and let $C_{i,j}$ be the result (time, number of variables or any other value) of the $i$-th method in the $j$-th benchmark. For every data set, we assign a number to every method: the best method in that data set has a 1, the second has a 2, and so on. Then, for every method, we compute the average of these values in the different data sets. The obtained value is denoted by $R_i$ and is called the *average rank* of the $i$-th method. A method with smaller average rank is better than a method with a bigger one.

These average ranks make it possible to rank the different methods. However, we are also interested in detecting whether the differences between the methods are significant or not: this is computed in the second part of the test. Before that, we need some previous definitions.

Given $i, j \in N = \{1, 2, \ldots, n\}$, we denote by $p_{i,j}$ the p-value[3] of $z_{i,j} = \frac{R_i - R_j}{\sqrt{n(n-1)/(6m)}}$ with respect a normal distribution $N(0, 1)$. A *partition* of $N = \{1, 2, \ldots, n\}$ is a collection of sets $P = \{P_1, P_2, \ldots, P_r\}$ such that (i) the $P_i$'s are subsets of $N$, (ii) $P_1 \cup P_2 \cup \cdots \cup P_r = N$ and

3. The p-value of $z$ with respect to a normal distribution $N(0, 1)$ is the probability $p[\ |Z| > |z|\ ]$, where the random variable $Z \sim N(0, 1)$.

(iii) $P_i \cap P_j = \emptyset$ for every $i \neq j$. Given $P$ a partition of $N$, we define

$$L(P) = \sum_{i=1}^{r} \frac{|P_i|(|P_i| - 1)}{2}$$

and $p(P)$ as the minimum $p_{i,j}$ such that $i$ and $j$ belong to the same subset $P_k \in P$.

The Bergmann-Hommel test ensures (with a significance level of $\alpha$) that the methods $i$ and $j$ are not significantly different if and only if there is a partition $P$ with $p(P) > \alpha L(P)$ such that $i$ and $j$ belong to the same subset $P_k \in P$. Hence, it is a time-consuming test since the number of partitions can be very large.

In our case, the data sets are the families of benchmarks. We have to use the families instead of the benchmarks because the data sets must be independent.

### 8.2 Encodings into SAT

We start by comparing different methods for encoding Pseudo-Boolean constraints into SAT. We have focused on the time spent by the encoding, the number of auxiliary variables used and the number of clauses. Moreover, for each benchmark family, we also report the number of PB-constraints that were encoded into SAT.

The encodings we have included in the experimental evaluation are: the adder encoding as presented by Eén and Sörensson (2006) (Adder), the consistent WatchDog encoding of Bailleux et al. (2009) (WD-1), its GAC version (WD-2), the encoding into ROBDDs without coefficient decomposition, using Algorithm 1 and the encoding from Section 6 (BDD-1); the encoding into ROBDDs after coefficient decomposition as explained in Section 4.2 (BDD-2), with Algorithm 1 and the encoding from Section 6; and the GAC approach from Section 4.3 (*BDD-3*), also with Algorithm 1 and the encoding from Section 6. Notice that BDD-1 method is very similar to the ROBDDs presented by Eén and Sörensson (2006). However, since Algorithm 1 produces every node only once, BDD-1 should be faster. Also, the encoding of Section 6 only creates two clauses per BDD node, as opposed to six clauses as suggested by Eén and Sörensson.

Table 2 shows the number of problems that the different methods could encode without timing out. The first column corresponds to the family of problems. The second column shows the number of problems in this family. The third and fourth columns contain the average number of SAT and Pseudo-Boolean constraints in the problem. For the experiments, we considered a constraint to be SAT if it is a clause or has at most 3 variables. Small PB constraints do not benefit from the above encodings and hence for these constraints a naive encoding into SAT was always used. The remaining columns correspond to the number of encoded problems without timing out. Time limit was set to 1800 seconds per benchmark.

Table 3 shows the time spent to encode the benchmarks by the different methods. As before, the first columns correspond to the family of problems, the number of problems in this family and the average number of SAT and Pseudo-Boolean constraints in the problems. The remaining columns correspond to the average encoding time (in seconds) per benchmarks of each method. Timeouts are counted as 1800 seconds in the average computation.

Table 4 shows the average number of auxiliary variables required for encoding the PB constraints (SAT constraints are not counted). The meaning of the first 4 columns is the same as before, and the others contain the average number of auxiliary variables (in thousands) of the benchmarks that did not time out.

| **Family** | **Pr** | **SAT** | **PB** | **Adder** | **WD-1** | **WD-2** | **BDD-1** | **BDD-2** | **BDD-3** |
|---|---|---|---|---|---|---|---|---|---|
| lopes | 200 | 502,671 | 592,715 | 188 | 188 | 118 | **197** | **197** | 188 |
| army | 12 | 192 | 451 | **12** | **12** | **12** | **12** | **12** | **12** |
| blast | 8 | 6,510 | 1,253 | **8** | **8** | **8** | **8** | **8** | **8** |
| cache | 9 | 181,100 | 4,507 | **9** | **9** | **9** | **9** | **9** | **9** |
| chnl | 21 | 0 | 125 | **21** | **21** | **21** | **21** | **21** | **21** |
| dbstv30 | 5 | 326,200 | 2,701 | **5** | **5** | 0 | **5** | **5** | 0 |
| dbstv40 | 5 | 985,200 | 4,801 | **5** | **5** | 0 | **5** | **5** | 0 |
| dbstv50 | 5 | 2,552,000 | 7501 | **5** | **5** | 0 | **5** | **5** | 0 |
| dlx | 3 | 20,907 | 857 | **3** | **3** | **3** | **3** | **3** | **3** |
| elf | 5 | 46,446 | 1,399 | **5** | **5** | **5** | **5** | **5** | **5** |
| fpga | 36 | 0 | 687 | **36** | **36** | **36** | **36** | **36** | **36** |
| j30 | 17 | 13,685 | 270 | **17** | **17** | **17** | **17** | **17** | **17** |
| j60 | 18 | 30,832 | 309 | **18** | **18** | **18** | **18** | **18** | **18** |
| j90 | 17 | 50,553 | 337 | **17** | **17** | 8 | **17** | **17** | 11 |
| j120 | 28 | 104,147 | 516 | **28** | **28** | 11 | **28** | **28** | 18 |
| neos | 4 | 1,451 | 3,831 | **4** | **4** | **4** | **4** | **4** | **4** |
| ooo | 19 | 95,217 | 4,487 | **19** | **19** | **19** | **19** | **19** | **19** |
| pig-crd | 20 | 0 | 113 | **20** | **20** | 18 | **20** | **20** | **20** |
| pig-cl | 20 | 161,150 | 58 | **20** | **20** | **20** | **20** | **20** | **20** |
| ppp | 6 | 29,846 | 1,023 | **6** | **6** | **6** | **6** | **6** | **6** |
| robin | 6 | 0 | 761 | **6** | **6** | 2 | **6** | **6** | **6** |
| 13queen | 100 | 8 | 93 | **100** | **100** | **100** | **100** | **100** | **100** |
| 11tsp11 | 100 | 2,662 | 45 | **100** | **100** | **100** | **100** | **100** | **100** |
| vdw | 5 | 8,978 | 267,840 | **5** | **5** | **5** | **5** | **5** | **5** |
| **TOTAL** | 669 | | | 657 | 657 | 540 | **666** | **666** | 626 |

Table 2: Number of problems encoded (without timing out) by the different methods.

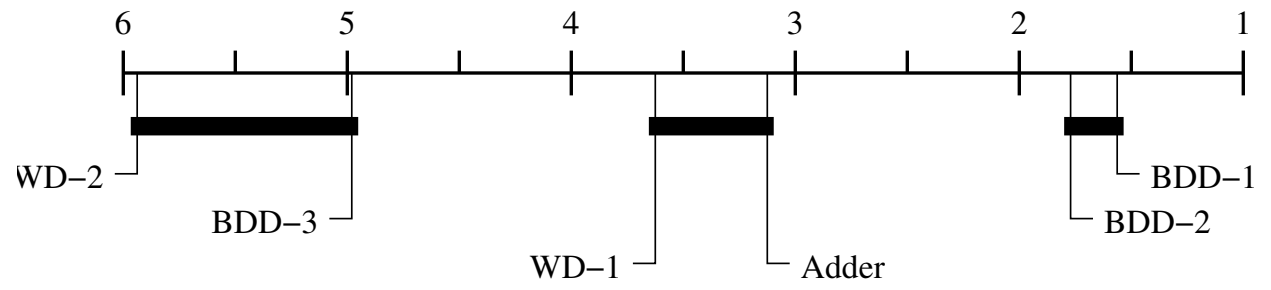

Figure 10: Statistical comparison of the results of Table 3, time spent by the different methods in encoding.

Finally, Table 5 contains the average number (in thousands) of clauses needed to encode the problem. As before, we have only considered the benchmarks that have not timed out, and clauses due to the encoding of SAT constraints are not counted.

Figures 10, 11 and 12 represent the Bergmann-Hommel tests of these tables. They show that BDD-1, BDD-2 and Adders are the best methods in terms of time, variables and clauses. It is worth mentioning that BDD-1 and BDD-2 are faster and use significantly less clauses than Adder. However, Adders uses significantly less auxiliary variables than BDD-2. Notice that BDD-1 is GAC, BDD-2 is only consistent and Adder is not consistent, so at least theoretically BDD-1 clauses have more unit propagation power than BDD-2 clauses, and BDD-2 clauses are better than Adder clauses. Hence, BDD-1 is the best method using these criteria and BDD-2 is better than Adder. Regarding the other methods, it seems clear

| Family | Pr | SAT | PB | Adder | WD-1 | WD-2 | BDD-1 | BDD-2 | BDD-3 |
|---|---|---|---|---|---|---|---|---|---|
| lopes | 200 | 502,671 | 592,715 | 335.23 | 292.14 | 996.07 | 165.75 | **163.66** | 316.35 |
| army | 12 | 192 | 451 | 0.37 | 0.43 | 39.98 | **0.19** | **0.19** | 10.26 |
| blast | 8 | 6,510 | 1,253 | 3.89 | 2.45 | 40.41 | 2.20 | **1.89** | 23.15 |
| cache | 9 | 181,100 | 4,507 | 23.08 | 18.74 | 81.65 | 16.19 | **15.74** | 47.78 |
| chnl | 21 | 0 | 125 | 0.54 | 1.05 | 87.08 | **0.13** | **0.13** | 2.68 |
| dbstv30 | 5 | 326,200 | 2,701 | 57.77 | 97.21 | — | **45.85** | 83.09 | — |
| dbstv40 | 5 | 985,200 | 4,801 | 211.51 | 210.25 | — | **105.62** | 165.96 | — |
| dbstv50 | 5 | 2,552,000 | 7,501 | 547.30 | 552.99 | — | **272.02** | 468.51 | — |
| dlx | 3 | 20,907 | 857 | 3.73 | 3.05 | 8.41 | 2.76 | **2.75** | 6.19 |
| elf | 5 | 46,446 | 1,399 | 7.37 | 6.53 | 21.68 | **5.19** | 5.90 | 13.42 |
| fpga | 36 | 0 | 687 | 1.90 | 2.46 | 69.90 | **0.30** | **0.30** | 3.75 |
| j30 | 17 | 13,685 | 270 | 3.64 | 4.62 | 81.03 | **3.13** | 3.67 | 42.44 |
| j60 | 18 | 30,832 | 309 | **6.85** | 10.69 | 466.07 | 8.19 | 8.77 | 252.69 |
| j90 | 17 | 50,553 | 337 | **14.81** | 31.02 | 1,277.28 | 28.20 | 27.76 | 1,155.18 |
| j120 | 28 | 104,147 | 516 | **19.25** | 47.62 | 1,305.55 | 21.68 | 25.50 | 967.10 |
| neos | 4 | 1,451 | 3,832 | 10.43 | 12.65 | 257.97 | **3.46** | 5.32 | 77.04 |
| ooo | 19 | 95,217 | 4,487 | 13.48 | 9.67 | 71.20 | **7.76** | 7.88 | 26.35 |
| pig-crd | 20 | 0 | 113 | 0.97 | 3.29 | 517.51 | 0.22 | **0.21** | 9.52 |
| pig-cl | 20 | 161,150 | 58 | 7.73 | 8.78 | 284.15 | 7.35 | **7.31** | 10.79 |
| ppp | 6 | 29,846 | 1,024 | 6.13 | 5.09 | 33.26 | **3.17** | 3.23 | 9.83 |
| robin | 6 | 0 | 761 | 12.03 | 67.41 | 1,315.96 | 2.94 | **2.82** | 301.11 |
| 13queen | 100 | 8 | 93 | 0.19 | 0.45 | 100.29 | **0.14** | 0.17 | 18.48 |
| 11tsp11 | 100 | 2,662 | 45 | 0.46 | 0.51 | 24.42 | **0.30** | 0.33 | 6.30 |
| vdw | 5 | 8,978 | 267,840 | 170.33 | 109.42 | 441.21 | 47.15 | **46.32** | 125.91 |
| **Average** | | | | 110.57 | 99.79 | 510.40 | **55.90** | 57.66 | 223.41 |

Table 3: Average time spent on the encoding by the different methods.

| Family | Pr | SAT | PB | Adder | WD-1 | WD-2 | BDD-1 | BDD-2 | BDD-3 |
|---|---|---|---|---|---|---|---|---|---|
| lopes | 200 | 502,671 | 592,715 | **1,744.05** | 3,566.11 | 5,478.81 | 2,393.97 | 2,394 | 7,734.65 |
| army | 12 | 192 | 451 | **4.63** | 10.96 | 245.49 | 6.36 | 6.36 | 479.83 |
| blast | 8 | 6,510 | 1,253 | **27.77** | 62.22 | 1,394.5 | 36.74 | 39.67 | 761.29 |
| cache | 9 | 181,100 | 4,507 | **145.66** | 339.18 | 2,393.97 | 201.18 | 210.83 | 1,503.19 |
| chnl | 21 | 0 | 125 | 8.39 | 24.55 | 1,007.9 | **6.76** | **6.76** | 184.59 |
| dbstv30 | 5 | 326,200 | 2,701 | **219.82** | 709.73 | — | 441.86 | 1,695.87 | — |
| dbstv40 | 5 | 985,200 | 4,801 | **2,468.45** | 6,564.44 | — | 4,282.16 | 7,225.63 | — |
| dbstv50 | 5 | 2,552,000 | 7,501 | **6,135.13** | 16,365.39 | — | 11,111.06 | 19,723.37 | — |
| dlx | 3 | 20,907 | 857 | **10.4** | 21.62 | 247.79 | 12.40 | 13.89 | 126.81 |
| elf | 5 | 46,446 | 1,399 | **20.37** | 42.78 | 571.38 | 24.62 | 28.13 | 306.76 |
| fpga | 36 | 0 | 687 | 21.15 | 53.96 | 1,074.03 | **13.27** | **13.27** | 242.03 |
| j30 | 17 | 13,685 | 270 | **18.15** | 50.8 | 1,190.59 | 44.96 | 59.82 | 1,153.51 |
| j60 | 18 | 30,832 | 309 | **37.03** | 112.35 | 4,775.72 | 157.92 | 180.05 | 7,285.69 |
| j90 | 17 | 50,553 | 337 | **65.4** | 217.01 | 6,543.52 | 553.8 | 553.76 | 19,793.49 |
| j120 | 28 | 104,147 | 516 | **159.43** | 540 | 5,713.75 | 612.13 | 806.08 | 22,246.82 |
| neos | 4 | 1,451 | 3,832 | **73.74** | 185.59 | 3,542.94 | 79.33 | 122.53 | 2,003.95 |
| ooo | 19 | 95,217 | 4,487 | **118.15** | 273.54 | 2,248.34 | 162.25 | 168.61 | 1,315.77 |
| pig-crd | 20 | 0 | 113 | 15.26 | 50.75 | 2,966.58 | **11.93** | **11.93** | 632.33 |
| pig-cl | 20 | 161,150 | 58 | 7.68 | 25.25 | 1,984.06 | **4.01** | **4.01** | 310.03 |
| ppp | 6 | 29,846 | 1,024 | **57.13** | 141.49 | 623.58 | 81.57 | 82.86 | 382.67 |
| robin | 6 | 0 | 761 | 171.67 | 628.45 | 3,634.13 | **158.55** | **158.55** | 16,565.75 |
| 13queen | 100 | 8 | 93 | **2.2** | 6.17 | 461.54 | 5.63 | 7.08 | 791.43 |
| 11tsp11 | 100 | 2,662 | 45 | **3.37** | 8.83 | 170.59 | 5.71 | 6.51 | 221.21 |
| vdw | 5 | 8,978 | 267,840 | 1,895.39 | 3,356.94 | 12,818.51 | **1,391.65** | **1,391.65** | 5,875.58 |
| **Average** | | | | **591.35** | 1,266.23 | 1,876.33 | 892.62 | 998.44 | 3,807.34 |

Table 4: Average number of auxiliary variables (in thousands) used.

| Family | Pr | SAT | PB | Adder | WD-1 | WD-2 | BDD-1 | BDD-2 | BDD-3 |
|---|---|---|---|---|---|---|---|---|---|
| lopes | 200 | 502,671 | 592,715 | 10,643.63 | 7,471.89 | 22,082.47 | 3,049.06 | **3,049.02** | 9,746.32 |
| army | 12 | 192 | 451 | 26.09 | 34.5 | 2,155.7 | **10.87** | **10.87** | 924.93 |
| blast | 8 | 6,510 | 1,253 | 184.95 | 102.07 | 2,108.48 | 70.9 | **65.46** | 1,264.83 |
| cache | 9 | 181,100 | 4,507 | 980.51 | 550.68 | 3,651.91 | **272.27** | 275.26 | 2,418.77 |
| chnl | 21 | 0 | 125 | 56.82 | 116.97 | 4,936.16 | **11.23** | **11.23** | 285.67 |
| dbstv30 | 5 | 326,200 | 2,701 | 1,497.41 | 3,367.75 | — | **857.39** | 3,282.22 | — |
| dbstv40 | 5 | 985,200 | 4,801 | 17,184.62 | 16,916.6 | — | **5,526.6** | 11,259.29 | — |
| dbstv50 | 5 | 2,552,000 | 7,501 | 42,797.38 | 44,310.83 | — | **14,400.14** | 31,279.2 | — |
| dlx | 3 | 20,907 | 857 | 65.25 | 35.68 | 377.76 | 23.04 | **22.59** | 208.92 |
| elf | 5 | 46,446 | 1,399 | 129.05 | 71.28 | 881.02 | 46.3 | **46.07** | 507.46 |
| fpga | 36 | 0 | 687 | 139.45 | 175.66 | 3,615.21 | **15.65** | **15.65** | 278.37 |
| j30 | 17 | 13,685 | 270 | 121.56 | 164.78 | 3,889.58 | **89.53** | 116.04 | 2,244.021 |
| j60 | 18 | 30,832 | 309 | **253.16** | 494.83 | 22,843.05 | 311.15 | 351.01 | 14,355.2 |
| j90 | 17 | 50,553 | 337 | **450.49** | 1,286.17 | 34,136.7 | 1,106.22 | 1,095.12 | 39,112.08 |
| j120 | 28 | 104,147 | 516 | **1,102.86** | 3,803.28 | 26,205.19 | 1,186.55 | 1,570.73 | 44,068.97 |
| neos | 4 | 1,451 | 3,832 | 471.47 | 594.72 | 12,410.1 | **139.26** | 220.45 | 3,681.26 |
| ooo | 19 | 95,217 | 4,487 | 793.36 | 442.47 | 3,378.16 | **219.41** | 227.61 | 2,126.42 |
| pig-crd | 20 | 0 | 113 | 104.68 | 367.03 | 20,711.11 | **19.86** | **19.86** | 958.65 |
| pig-cl | 20 | 161,150 | 58 | 52.41 | 180.33 | 14,641.26 | **4.07** | **4.07** | 314.05 |
| ppp | 6 | 29,846 | 1,024 | 392.6 | 271.66 | 1,804.29 | **100.89** | 103.24 | 654.48 |
| robin | 6 | 0 | 761 | 1,185.92 | 6,916.35 | 19,694.86 | **281.42** | **281.42** | 28,875.61 |
| 13queen | 100 | 8 | 93 | 14.73 | 38.91 | 5,068.35 | **10.84** | 13.73 | 1,573.89 |
| 11tsp11 | 100 | 2,662 | 45 | 23.31 | 25.35 | 1,335.6 | **7.76** | 9.35 | 433.59 |
| vdw | 5 | 8,978 | 267,840 | 10,885.96 | 6,564.14 | 24,274.21 | **1,662.73** | **1,662.73** | 7,262.88 |
| **Average** | | | | 3,675.7 | 2,970.54 | 8,297.1 | 1,174.38 | 1,380.41 | 5,843.53 |

Table 5: Average number of clauses (in thousands) used.

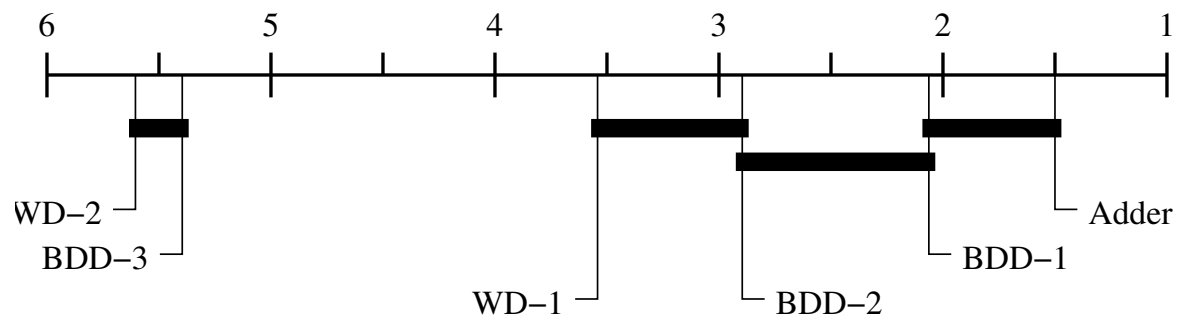

Figure 11: Statistical comparison of the results of Table 4, number of auxiliary variables used by the different encodings.

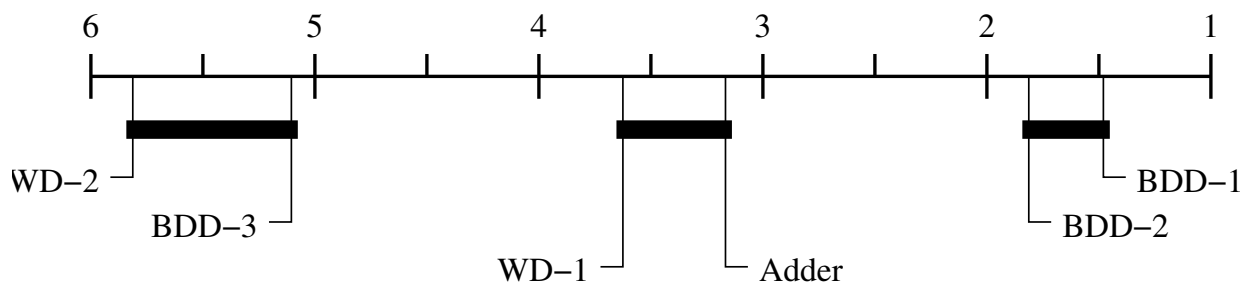

Figure 12: Statistical comparison of the results of Table 5, number of clauses used by the different methods.

| **Family** | **Adder** | **WD-1** | **WD-2** | **BDD-1** | **BDD-2** | **BDD-3** | **bsolo** | **MiniSAT** | **SAT4J** | **Wbo** | **borg** | **SMT** | **VBS** |
|---|---|---|---|---|---|---|---|---|---|---|---|---|---|
| lopes | 42 | 54 | 40 | 56 | 57 | 61 | 39 | **66** | 23 | 63 | 37 | 43 | 77 |
| army | 9 | **12** | 7 | 10 | 11 | 5 | 6 | 6 | 6 | 6 | 10 | 5 | 12 |
| blast | **8** | **8** | **8** | **8** | **8** | **8** | **8** | **8** | **8** | **8** | **8** | **8** | 8 |
| cache | **9** | **9** | **9** | **9** | **9** | **9** | 7 | 8 | 6 | 6 | 6 | **9** | 9 |
| chnl | 3 | 3 | 2 | 5 | 5 | 3 | **21** | 3 | 1 | 3 | **21** | 0 | 21 |
| dbstv30 | **5** | **5** | 0 | **5** | **5** | 0 | **5** | **5** | **5** | **5** | **5** | **5** | 5 |
| dbstv40 | 0 | **5** | 0 | **5** | **5** | 0 | **5** | **5** | **5** | **5** | **5** | **5** | 5 |
| dbstv50 | 0 | **5** | 0 | **5** | **5** | 0 | **5** | **5** | **5** | **5** | **5** | **5** | 5 |
| dlx | **3** | **3** | **3** | **3** | **3** | **3** | **3** | **3** | **3** | **3** | **3** | **3** | 3 |
| elf | **5** | **5** | **5** | **5** | **5** | **5** | **5** | **5** | **5** | **5** | **5** | **5** | 5 |
| fpga | 25 | **36** | **36** | **36** | **36** | **36** | **36** | 33 | **36** | **36** | **36** | **36** | 36 |
| j30 | **17** | **17** | **17** | **17** | **17** | **17** | **17** | **17** | **17** | **17** | **17** | **17** | 17 |
| j60 | **17** | **17** | **17** | **17** | **17** | **17** | **17** | **17** | **17** | **17** | **17** | **17** | 17 |
| j90 | **17** | **17** | 7 | **17** | **17** | 8 | **17** | **17** | **17** | **17** | **17** | **17** | 17 |
| j120 | 14 | **16** | 9 | **16** | **16** | 11 | 13 | 12 | **16** | **16** | **16** | **16** | 17 |
| neos | **2** | **2** | **2** | **2** | **2** | **2** | **2** | **2** | **2** | **2** | **2** | **2** | 2 |
| ooo | 15 | **19** | 16 | 18 | **19** | 17 | 14 | 15 | 14 | 15 | 14 | 17 | 19 |
| pig-crd | 2 | 2 | 2 | 2 | 2 | 1 | 19 | 2 | 2 | 2 | **20** | 0 | 20 |
| pig-cl | 2 | 1 | 2 | 1 | 1 | 2 | 3 | 2 | 2 | 2 | **5** | 0 | 5 |
| ppp | 4 | 3 | 4 | 3 | 4 | 4 | 4 | 4 | 4 | 3 | **5** | 4 | 6 |
| robin | 3 | 3 | 2 | 3 | 3 | **6** | 3 | 3 | 4 | 3 | 3 | 4 | 6 |
| 13queen | **100** | **100** | **100** | **100** | **100** | **100** | **100** | **100** | **100** | **100** | **100** | **100** | 100 |
| 11tsp11 | **100** | **100** | 96 | **100** | **100** | 75 | 72 | 90 | 93 | **100** | **100** | **100** | 100 |
| vdw | **1** | **1** | **1** | **1** | **1** | **1** | **1** | **1** | **1** | **1** | **1** | **1** | 2 |
| **TOTAL** | 403 | 443 | 385 | 444 | 448 | 391 | 422 | 429 | 392 | 440 | **458** | 419 | 514 |

Table 6: Number of problems solved by different methods.

that encoding $n$ different constraints in order to obtain GAC, as it is done in WD-2 and BDD-3, is not a good idea in terms of variables and clauses.

### 8.3 SAT vs. PB

In this section we compare the state-of-the-art solvers for Pseudo-Boolean problems and some encodings into SAT. For the SAT approach, once the encoding has been done, the SAT formula is given to the SAT Solver Lingeling (Biere, 2010) version 276. We have considered the same SAT encodings as in the previous section. Regarding Pseudo-Boolean solvers, we have considered MiniSAT+ (Eén & Sörensson, 2006) and the best non-parallel solvers in the *No optimization, small integers, linear constraints* category of the Pseudo-Boolean Competition: borg (Silverthorn & Miikkulainen, 2010) version pb-dec-11.04.03, bsolo (Manquinho & Silva, 2006) version 3.2, wbo (Manquinho, Martins, & Lynce, 2010) version 1.4 and SAT4J (Berre & Parrain, 2010) version 2.2.1. We have also included the SMT Solver Barcelogic (Bofill, Nieuwenhuis, Oliveras, Rodríguez-Carbonell, & Rubio, 2008) for PB constraints, which couples a SAT solver with a theory solver for PB constraints.

Table 6 shows the number of instances solved by each method. Table 7 shows the average time spent by all these methods. For the SAT encodings, times include both the encoding and SAT solving time. As before, a time limit of 1800 seconds per benchmark was set, and for the average computation, a timeout is counted as 1800 seconds. Both tables include a column VBS (Virtual Best Solver), which represents the best solver in every instance. This gives an idea of which speedup we could obtain with a portfolio approach.

Figure 13 shows the result of the Bergmann-Hommel test: SMT is the best method, whereas Adder, BDD-3 and WD-2 are the worst ones. There are no significant difference between the other methods. The main conclusion we can infer is that BDD encodings are definitely a competitive method. Also, there is no technique that outperforms the others in all benchmark families, and hence portfolio strategies would make a lot of sense in this area, as witnessed by the performance of Borg, which implements such an approach. Finally, we also want to mention that the possible exponential explosion of BDDs rarely occurs in practice and hence, coefficient decomposition does not seem to pay off in practical situations.

| **Family** | **Adder** | **WD-1** | **WD-2** | **BDD-1** | **BDD-2** | **BDD-3** | **bsolo** | **MiniSAT** | **SAT4J** | **Wbo** | **borg** | **SMT** | **VBS** |
|---|---|---|---|---|---|---|---|---|---|---|---|---|---|
| lopes | 1,515 | 1,420 | 1,561 | 1,408 | 1,401 | 1,435 | 1,509 | **1,344** | 1,661 | 1,364 | 1,555 | 1,464 | 1,249 |
| army | 660 | **139** | 1,141 | 543 | 469 | 1,298 | 1,028 | 913 | 1,127 | 1,084 | 438 | 1,066 | 86 |
| blast | 6.12 | 2.56 | 46.78 | 2.42 | 1.99 | 27.63 | 0.12 | 0.51 | 0.84 | 0.08 | 2.13 | **0.03** | 0.03 |
| cache | 253 | 123 | 396 | **75.49** | 115 | 375 | 653 | 395 | 670 | 606 | 636 | 266 | 63.95 |
| chnl | 1,543 | 1,543 | 1,716 | 1,508 | 1,508 | 1,681 | **0.55** | 1,551 | 1,751 | 1,673 | 3.78 | — | 0.47 |
| dbstv30 | 1,049 | 128 | — | 91.66 | 192 | — | 59.28 | 32.6 | 99.81 | 1.54 | 9.87 | **1.28** | 1.28 |
| dbstv40 | — | 366 | — | 198 | 324 | — | 187 | 72.25 | 9.74 | 5.69 | 45.33 | **4.44** | 4.44 |
| dbstv50 | — | 935 | — | 629 | 792 | — | 200 | 430 | 21.22 | 16.13 | 121 | **11.36** | 11.36 |
| dlx | 7.06 | 4.88 | 25.72 | 4.29 | 4.34 | 19.58 | 3.47 | 1.29 | 1.6 | 0.55 | 3.15 | **0.17** | 0.17 |
| elf | 13.87 | 10.14 | 44.09 | 7.97 | 9 | 30.03 | 28.58 | 2.97 | 2.31 | 1.42 | 11.61 | **0.69** | 0.69 |
| fpga | 586 | 5.27 | 113 | 0.92 | 0.92 | 37.64 | 0.27 | 242 | 1.47 | 5.17 | 3.04 | **0.1** | 0.07 |
| j30 | 16.7 | 7.79 | 116 | 5.94 | 8.42 | 77.88 | 6.53 | 4.6 | 14.57 | 0.53 | 1.93 | **0.28** | 0.28 |
| j60 | 137 | 114 | 551 | 113 | 116 | 398 | 110 | 115 | 105 | 101 | 104 | **101** | 101 |
| j90 | 24.18 | 36.63 | 1,303 | 39.72 | 39.46 | 1,233 | 0.9 | 3.96 | 1.42 | 0.41 | 3.32 | **0.15** | 0.15 |
| j120 | 978 | 854 | 1,364 | 839 | 851 | 1,262 | 967 | 1,031 | 849 | 839 | 841 | **814** | 756 |
| neos | 1,023 | 936 | 1,405 | 910 | 915 | 1,073 | 1,106 | 1,276 | 1,038 | **901** | 976 | 925 | 901 |
| ooo | 479 | 190 | 493 | **151** | 176 | 488 | 645 | 453 | 575 | 486 | 512 | 259 | 126 |
| pig-crd | 1,620 | 1,620 | 1,680 | 1,620 | 1,620 | 1,725 | 114 | 1,626 | 1,749 | 1,685 | **3.92** | — | 1.92 |
| pig-cl | 1,624 | 1,715 | 1,693 | 1,718 | 1,718 | 1,721 | 1,658 | 1,623 | 1,705 | 1,742 | **1,369** | — | 1,367 |
| ppp | 631 | 1,001 | 656 | 906 | 858 | 646 | 605 | 919 | 602 | 901 | **390** | 601 | 210 |
| robin | 938 | 921 | 1,353 | 913 | 913 | 719 | 936 | 971 | 778 | 920 | 963 | **605** | 444 |
| 13queen | 47.52 | **1.64** | 264 | 4.63 | 4.51 | 643 | 54.82 | 238 | 18.92 | 5.9 | 20.35 | 1.92 | 1.28 |
| 11tsp11 | 28.36 | 8.29 | 428.6 | 23.86 | 18.32 | 731 | 855 | 369 | 503 | 229 | 27.64 | **1.81** | 1.51 |
| vdw | 1,645 | 1,568 | 1,545 | 1,493 | 1,493 | 1,612 | 1,478 | 1,448 | 1,596 | 1,441 | 1,450 | **1,441** | 1,186 |
| **Av.** | 783 | 669 | 958 | 667 | 667 | 1,003 | 764 | 772 | 849 | 710 | **613** | 696 | 475 |

Table 7: Time spent by different methods on solving the problem (in seconds).

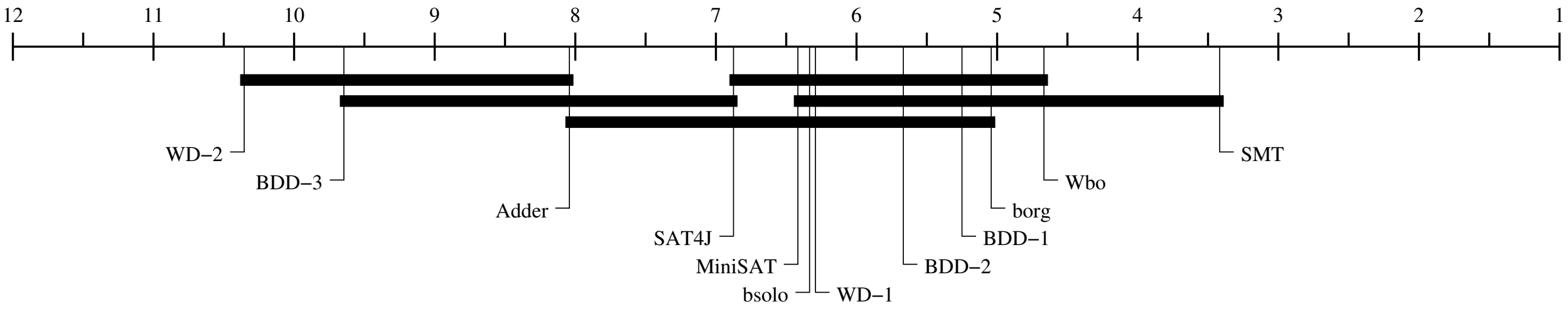

Figure 13: Statistical comparison of the results of Table 7, runtime of the different methods.

Regarding the Best Virtual Solver, SMT contributes to 52% of the problems. In 25% of the cases the best solution was given by a specific PB solver. Among them, Wbo contributes with 10% of the problems and bsolo with 8%. Finally, encoding methods give the best solution in the 23% of the cases: 14% of the times due to Watchdog methods and 8% of the times due to BDD-based methods.

### 8.4 Sharing

One of the possible advantages of using ROBDDs to encode Pseudo-Boolean constraints is that ROBDDs allow one to encode a set of constraints, and not just one. It would seem natural to think that if two constraints are *similar enough*, the two individual ROBDDs would be similar in structure, and merging them into a single one would result in a ROBDD whose size is smaller than the sum of the two individual ROBDDs. However, the main difficulty is to decide which constraints should be encoded together, since a bad choice could result in a ROBDD whose size is larger than the sum of the ROBDDs for the individual constraints.

We performed initial experiments where the criteria of similarity between constraints only took into account which variables appeared in the constraints. We first fixed an integer $k$ and chose the constraint with the largest set of variables. After that, we looked for a constraint such that all but $k$ variables appeared in the first constraint. The next step was to look for another constraint such that all but $k$ variables appeared in any of the two previous constraints and so on, until reaching a fixpoint. Finally, all selected constraints were encoded together.

We tried this experiment on all benchmarks with different values of $k$ and it rarely gave any advantage. However, we still believe that there could be a way of encoding different constraints into a single ROBDD, but different criteria for selecting the constraints should be studied. We see this as a possible line of future research.

## 9. Conclusions and Future Work

Both theoretical and practical contributions have been made. Regarding the theoretical part, we have negatively answered the question of whether all PB constraints admit polynomial BDDs by citing the work of Hosaka et al. (1994) which, to the best of our knowledge, is largely unknown in our research community. Moreover, we have given a simpler proof assuming that NP is different from co-NP, which relates the subset sum problem and the ROBDDs' size of PB constraints.

At the practical level, we have introduced a ROBDD-based polynomial and generalized arc-consistent encoding of PB constraints and developed a BDD-based generalized arc-consistent encoding of monotonic functions that only uses two clauses per BDD node. We have also presented an algorithm to efficiently construct all these ROBDDs and proved that the overall method is competitive in practice with state-of-the-art encodings and tools. As future work at the practical level, we plan to study which type of Pseudo-Boolean constraints are likely to produce smaller ROBDDs if encoded together rather than being encoded individually.

## Acknowledgments

All UPC authors are partially supported by Spanish Min. of Educ. and Science through the LogicTools-2 project (TIN2007-68093-C02-01). Abío is also partially supported by FPU grant.